\documentclass{article}

\usepackage[final]{corl_2019} 
\usepackage{amsmath}
\usepackage{ dsfont }
\usepackage{algorithmic}
\usepackage{algorithm}
\usepackage{graphicx}
\usepackage{subcaption}
\usepackage{amssymb}

\usepackage{cleveref}

\usepackage{wrapfig}

\usepackage{microtype}

\usepackage{enumitem}

\newcommand{\fig}[1]{Fig.~\ref{#1}}
\newcommand{\eq}[1]{Eq.~\eqref{#1}}

\newcommand{\design}{\ensuremath{\xi}}
\newcommand{\designspace}{\ensuremath{{\Xi}}}

\newcommand{\klstate}{\ensuremath{\mathbf{\text{s}}}}
\newcommand{\klaction}{\ensuremath{\mathbf{\text{a}}}}
\newcommand{\klpolicy}{\ensuremath{\pi}}
\newcommand{\klreward}{\ensuremath{\text{r}}}
\newcommand{\klq}{\ensuremath{Q}}
\newcommand{\kllocal}{\ensuremath{\text{Ind.}}}
\newcommand{\klglobal}{\ensuremath{\text{Pop.}}}

\newcommand{\klreplayglobal}{\ensuremath{\text{Replay}_{\klglobal}}}
\newcommand{\klreplaylocal}{\ensuremath{\text{Replay}_{\kllocal}}}
\newcommand{\klreplaystart}{\ensuremath{\text{Replay}_{\klstate_0}}}

\newcommand{\klexpect}[1]{\ensuremath{\mathds{E}\left[ #1 \right]}}
\newcommand{\klexpectdist}[2]{\ensuremath{\mathds{E}_{#2}\left[ #1 \right]}}

\title{Data-efficient Co-Adaptation of Morphology and Behaviour with Deep Reinforcement Learning}

\newcommand{\klalgname}{Fast Evolution through Actor-Critic Reinforcement Learning}

\author{
  Kevin Sebastian Luck\\
  Interactive Robotics Lab\\
  Arizona State University\\
  United States\\
  \texttt{ksluck@asu.edu} \\
  \And
  Heni Ben Amor\\
  Interactive Robotics Lab\\
  Arizona State University\\ 
  United States\\
  \texttt{hbenamor@asu.edu} \\
  \And
  Roberto Calandra\\
  Facebook AI Research\\
  United States\\
  \texttt{rcalandra@fb.com} \\
}

\begin{document}
\maketitle


\begin{abstract}
Humans and animals are capable of quickly learning new behaviours to solve new tasks. Yet, we often forget that they also rely on a highly specialized morphology that co-adapted with motor control throughout thousands of years.
Although compelling, the idea of co-adapting morphology and behaviours in robots is often unfeasible because of the long manufacturing times, and the need to re-design an appropriate controller for each morphology.
In this paper, we propose a novel approach to automatically and efficiently co-adapt a robot morphology and its controller.
Our approach is based on recent advances in deep reinforcement learning, and specifically the soft actor critic algorithm.
Key to our approach is the possibility of leveraging previously tested morphologies and behaviors to estimate the performance of new candidate morphologies.
As such, we can make full use of the information available for making more informed decisions, with the ultimate goal of achieving a more data-efficient co-adaptation (i.e., reducing the number of morphologies and behaviors tested).
Simulated experiments show that our approach requires drastically less design prototypes to find good morphology-behaviour combinations, making this method particularly suitable for future co-adaptation of robot designs in the real world.

\end{abstract}


\section{Introduction}

\begin{figure}[t!]
    \centering
    \begin{subfigure}[b]{0.49\textwidth}
        \includegraphics[width=\textwidth]{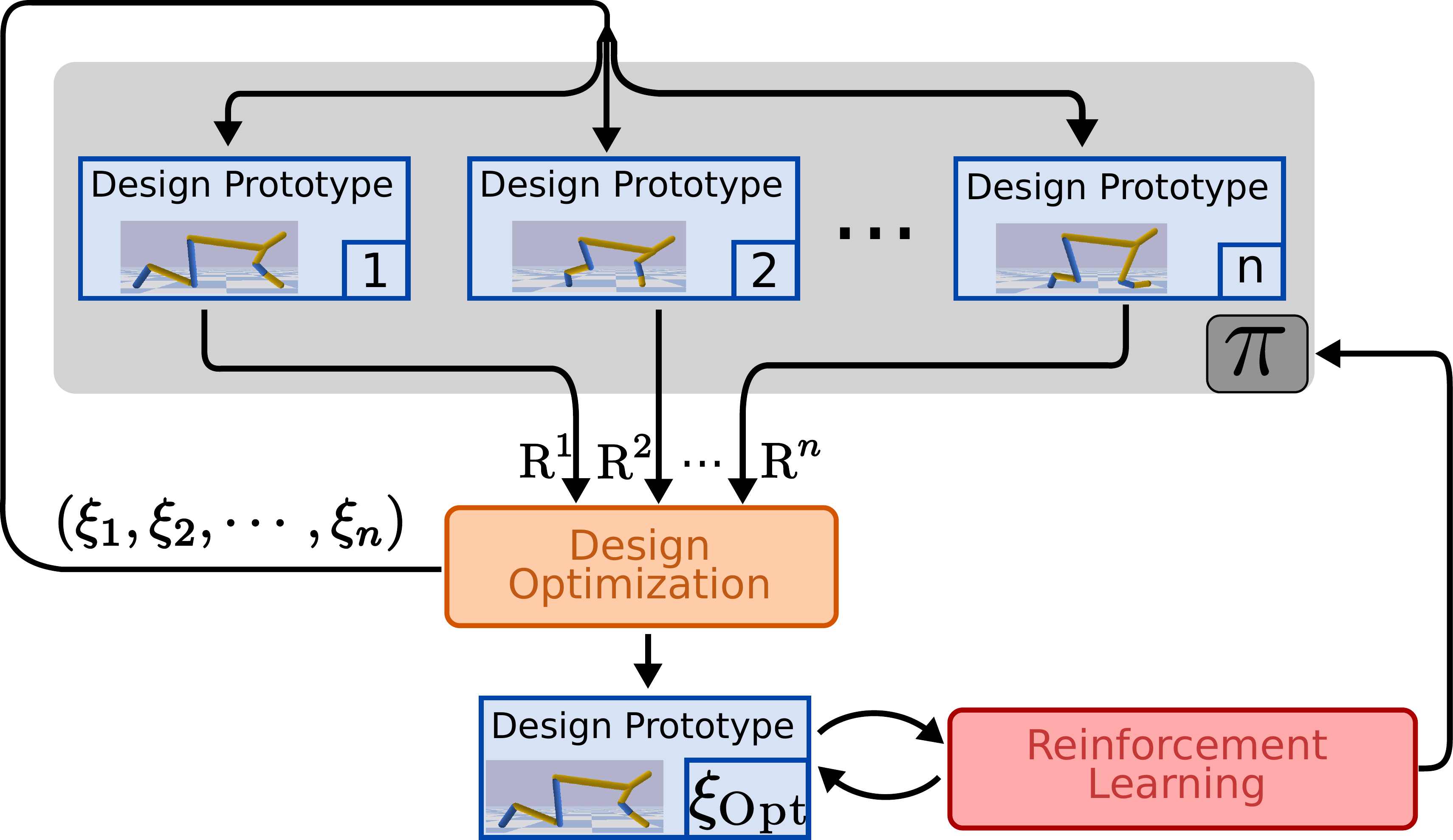}
        \caption{Traditional approach in which designs are evaluated in simulation during the design optimization process.}
        \label{Fig::concepts::normal}
    \end{subfigure}
    \hfill
    \begin{subfigure}[b]{0.49\textwidth}
        \includegraphics[width=\textwidth]{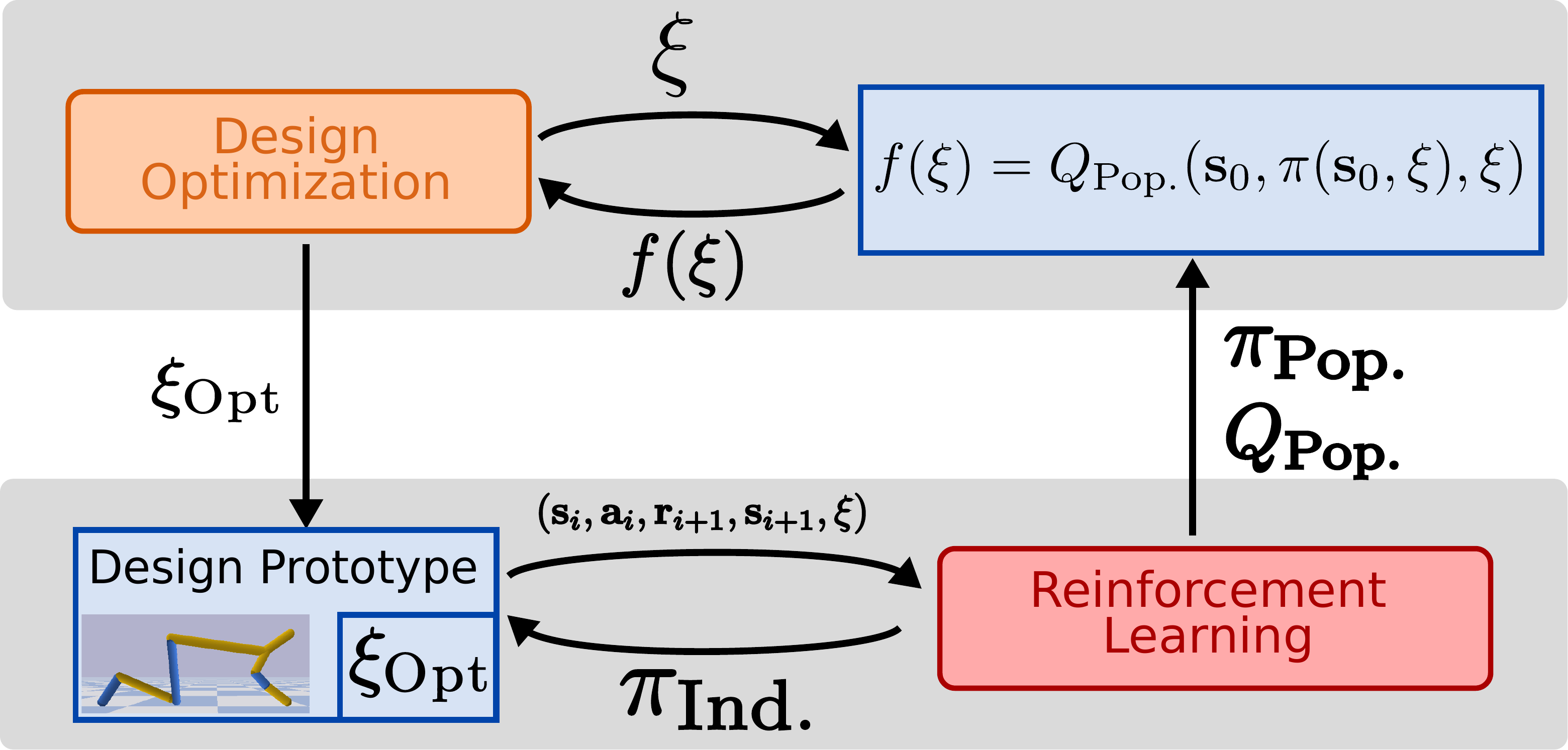}
        \caption{Proposed method which evaluates designs with the learned Q- and policy network and thus reduces the amount of simulations or physical prototypes required during the design optimization process.}
        \label{Fig::concepts::proposed}
    \end{subfigure}
    
    \vspace{0.35cm}
    \begin{subfigure}[b]{0.99\textwidth}
        \centering
        \includegraphics[width=0.99\textwidth]{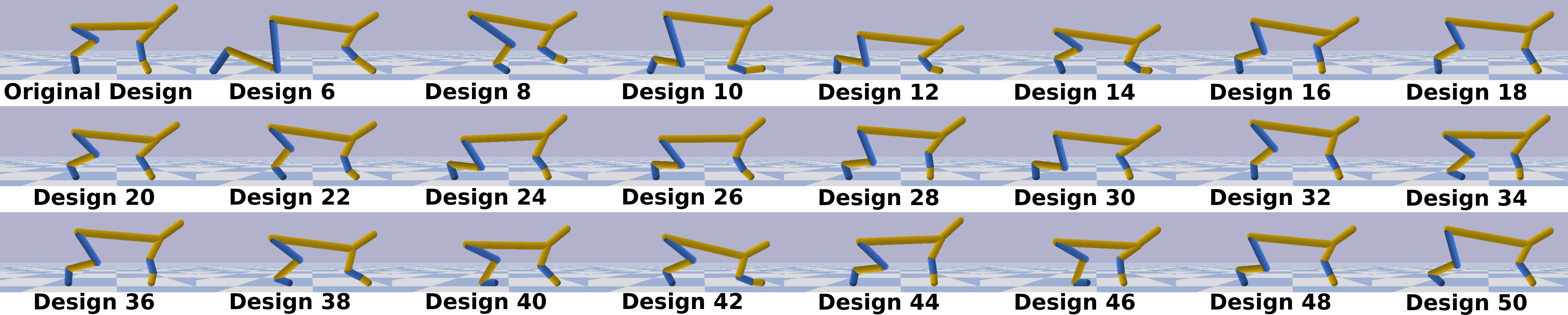}
        \caption{Designs $\design_{\text{Opt}}$ selected by the proposed method for the Half-Cheetah task.}
    \end{subfigure}
    
    \caption{
    We propose to (b) use an actor and critic for design exploration instead of (a) creating design prototypes and evaluating their performance in simulation or the real world. 
    Our goal is to reduce the amount of time needed to (c) evolve a robotic prototype in the real world.}
    \label{Fig::concepts}
\end{figure}

In nature, both morphology and behaviour of a species crucially shape its physical interactions with the environment \cite{bertossa2011evodevo}. 
For example, the diversity in animal locomotion styles is an immediate result of the interplay between different body structures, e.g., different numbers, compositions and shapes of limbs, as well as as different neuromuscular controls, e.g., different sensory-motor loops and neural periodic patterns. 
Adaptation of a species to new ecological opportunities often comes with changes to both body shape and control signals -- \textit{morphology and behaviour are co-adapted}. 
Building upon this insight, we investigate in this paper a methodology for co-adaptation of the morphology and behaviour for computational agents using deep reinforcement learning. 
Without loss of generality, we focus in particular on legged locomotion. 
The goal of legged robots in such locomotion tasks is to transform as much electric energy as possible into directional movement~\cite{nygaard2018real, seok2014design, alexander1984walking, jansen2017bio}. 
To this end, two approaches exist: 1) optimization of the behavioural policy, and 2) optimization of the robot design, which affects the achievable locomotion efficiency \cite{sims1994evolving, schaff2018jointly, nygaard2018real, luck2017lab}. 
Policy optimization is, especially in novel or changing environments, often performed using reinforcement learning~\cite{deisenroth2013survey, luck2017lab}. 
Design optimization is frequently based on evolutionary algorithms or evolution-inspired and use a population of design prototypes for this process (Fig. \ref{Fig::concepts::normal}) \cite{sims1994evolving, nygaard2018real, corucci2015novelty}. 
However, manufacturing and evaluating a large quantity of design candidates is often infeasible in the real world due to cost and time constraints, especially for larger robots. 
Therefore, the evaluation of designs is often restricted to simulation, which is feasible but suffers from the simulation-to-reality-gap \cite{zagal2004back, koos2012transferability}. 
Designs and control policies optimized in simulation are often not the best possible choice for the real world, especially if the robotics system is complex and the environmental parameters hard to model. 
For example, in the work of Lipson and Pollack \cite{lipson2000automatic} designs were first optimized in simulation in an evolutionary manner and then manufactured in the real world. 
However, the performances of the manufactured designs in the real world were significant lower than in simulation in all but one case (see Table 1 in \cite{lipson2000automatic}), even though efforts were undertaken to close the simulation-to-reality gap for the described robot.

The method proposed in this work caters towards the need of roboticists for data-efficiency in respect to the number of prototypes required to achieve an optimal design. 
We are combining design optimization and reinforcement learning in such a way that the reinforcement learning process provides us with an objective function for the design optimization process (Fig. \ref{Fig::concepts::proposed}). 
Thus, eliminating the need for a population of prototypes and requiring only one functioning prototype at a time. 

\section{Related Work}

The work of Schaff et al. \cite{schaff2018jointly} is a relatively recent approach to combine reinforcement learning and design optimization into one framework. 
The common idea is to consider the design parameter~$\design$ as an additional input to the policy $\klpolicy(\klstate, \design)$ and to optimize the expected reward~$\klexpect{R}$ given the policy and design. 
The policy is trained such that it is able to generalize over many designs and is iteratively updated with experience collected from a population of $n$ prototypes. 
The algorithm maintains a distribution over designs, whose parameters are optimized to maximize the expected reward.  However, this approach \cite{schaff2018jointly} requires the maintenance of a population of designs, which is updated every $t$ timesteps and relies on the simulator to compute the fitness of designs. 
Similarly, the work of David Ha \cite{ha2018reinforcement} uses the design parameters~$\design$ as input to the policy~$\klpolicy(\klstate, \design)$ but uses REINFORCE \cite{williams1992simple} to update the design parameters. 
Again, this approach requires a population of design prototypes to compute the introduced population-based policy gradient for the design as well as rewards collected from the simulator. 
The recent method introduced by \citet{liao2019data} employs Batch Bayesian Optimization to improve morphology and policies. 
The expected performance of designs is here learned and inferred by Gaussian Processes (GP), a second GP is also used to optimize the parameters of central pattern generators representing movement policies. 
The paper demonstrates the design optimization of a simulated micro-robot with three parameters defining the morphology. 
While the presented results are using a prototype population of 5 designs, the authors mention that the proposed method can handle a single prototype as well. 
One drawback of \cite{liao2019data} is, however, that the GP predicting the fitness of designs is trained only with a single value per design: the single highest reward achieved for a design. 
Since the maximum reward is potentially affected by the initial state a robot is in, this approach has a reduced applicability to tasks with noisy or random start states. 
In \cite{nygaard2018real}, the leg lengths and controller of a quadruped robot were optimized in the real world. 
The controller was here based on the inverse kinematics of the robot and defined by tuning eight parameters. 
All leg segment lengths were described by a two-dimensional design vector. 
Two different evolutionary algorithms were used to optimize these parameters over eight generations with a population size of eight and based on the reward received. 
While this experiment is an impressive demonstration of the potential of adapting behaviour and morphology in the real world, the task was simplified through the use of a re-configurable robot which is able to adapt its leg-lengths automatically. 
This decreases the setup-time required between experiments because manufacturing of leg-segments or other body parts are not necessary. 
All four of these approaches rely on a population of design prototypes whose performance must be evaluated in simulation or the real world, or rely on a single reward.


\section{Problem Statement} 
We formalize the problem of co-adapting morphology and behavior as the optimization 
\begin{align}
\theta^* = \text{arg max}_\theta\, R|_\theta\,,
\end{align}
of the reward $R$ w.r.t. the variables $\theta = [\design, \klpolicy]$ were $\design$ are the morphological properties of the agent, and $\klpolicy$ the behavior.
There are multiple ways to tackle this problem.
One commonly used way is to decompose it as bi-level optimization, where we iteratively optimize the morphology first $\design$, and after fixing it, we optimize the behavior~$\klpolicy$.
One advantage of this formulation is that by decoupling the two optimization, we can take into consideration the fact that evaluating different morphologies has an associated cost (e.g., manufacturing a physical robot) which can be substantially higher than evaluating different behaviors (e.g., running multiple controllers).
In this paper, we frame the learning of the behaviors as an extension of the standard Markov decision process (MDP)~\citep{Bellman1957} given the additional design variable~$\design$ (i.e., the context).
In this model, the transition probability to reach a state~$\klstate_{t+1}$ after performing action~$\klaction_t$ is given by $p(\klstate_{t+1} \vert \klstate_t, \klaction_t, \design)$ and depends on design properties $\design$ of the agent.  
The reward function $\klreward(\klstate, \klaction, \design)$ can be dependent on the design as well. 
For notational clarity, we will generally use $r(\klstate)$ in the remainder of the paper. 
The actions are generated from the policy $\klpolicy(\klstate, \design)$ and the goal is to maximize the expected future reward given by
\begin{equation}
    \klexpectdist{\sum_{i=0}^{\infty} \gamma^i \klreward(\klstate_{t+i+1}, \klaction_{t+i+1}, \design) \middle\vert \klstate_t = \klstate, \klaction_i = \klpolicy(\klstate_i), \design}{\klpolicy},
\end{equation}
with $\gamma \in [0,1]$ being a discount factor and future states $\klstate_{t+i+1}$ produced by the transition function. 
Our goal is hence to maximize this objective function for both the policy $\klpolicy$ and the design $\design$ using deep reinforcement learning.


\section{Optimization of Morphology and Behaviour}
We now introduce our proposed framework for sample-efficient optimization of behaviour and design for robotic prototypes. 
We first describe our novel objective function based on an actor and critic to remove the dependency on prototypes and simulations during design optimization. 
Thereafter, a method is described for fast behaviour adaptation by training a copy of actor and critic primarily on experience collected with the current design prototype. 
We continue with an explanation of two different design exploration mechanisms, random selection and novelty search. 
The chapter closes with a description of the reinforcement learning algorithms and optimization routines used.

\subsection{Using the Q-Function for Design Optimization}
Optimizing the behaviour of an agent usually requires learning a value or Q-value function and a policy $\klpolicy$ by the means of reinforcement learning. The rationale of our approach is to extend this methodology to the  evaluation of the space of designs, thereby reducing the need for large numbers of simulations or manufactured robot prototypes. 

The goal of \textit{design optimization} is to increase the efficiency of the agent given an optimal policy for each design. 
The objective function for this case can be the sum of rewards collected by evaluating the behaviour of the agent with this design, given by
\begin{equation}
    \max_{\design}\, \klexpectdist{\sum_{i=0}^T \klreward_t}{\klpolicy},
\end{equation}
where the rewards are collected through the execution of a policy $\klpolicy$ on the agent with design $\design$ in the real world or in simulation. 

To alleviate the aforementioned problems with the evaluation through executions in simulation or real world,
we instead propose to reuse the Q-function learned by a deep reinforcement learning algorithm and re-formulate our objective as 
\begin{equation}
    \max_{\design}\, \klexpectdist{\klq(\klstate, \klaction, \design)\middle\vert \klaction = \klpolicy(\klstate, \design)}{\klpolicy}, 
    \label{Eq::opt_2}
\end{equation}
where the action $\klaction$ is given by the policy $\klpolicy(\klstate, \design)$. 
This creates a strong coupling between the design optimization and reinforcement learning loop: 
We effectively reduce the problem of finding optimal designs to the problem of training a critic which is able to generate an estimated performance of a design given state and action. 
This means, while optimizing a policy for a design, we also train the objective function given above at the same time. We hypothesize that, during the training process, the critic learns to distinguish and interpolate between designs due to the influence of the design on the reward of transitions. 
We further reformulate Eq. \ref{Eq::opt_2} to optimize over the distribution of start states encountered in trajectories $(\klstate_0, \klaction_0, \klstate_1, \cdots, \klstate_T)$. 
The objective function becomes then the expected future reward given a design choice $\design$. 
This could be, for example, the case if the leg lengths of a robot are optimized and the initial position is a standing one. 
Here, the initial height of the robot would vary with the design choice. 
Thus, we reformulate the objective function in Eq. \ref{Eq::opt_2} such that we optimize over the distribution of start states with
\begin{equation}
    \max_{\design}\, \klexpectdist{\klexpectdist{\klq(\klstate_0, \klaction_0, \design)\middle\vert \klaction_0 = \klpolicy(\klstate_0, \design)}{\klpolicy}}{\klstate_0 \sim p(\klstate_0 \vert \design)}.
    \label{Eq::opt_3}
\end{equation}
The motivation to optimize this function over the distribution of start states is to take potential randomness in the initial positions, or even inaccuracies when resetting the initial position of a robot, into account. 
Since the distribution of start states might be unknown or even depend on the design, we approximate the expectation by drawing a random batch of start states $\klstate_0$ from a replay buffer, which contains exclusively all start states seen so far. 
If we use a deterministic deep neural network for policy $\klpolicy$, Eq. \ref{Eq::opt_3} reduces to
\begin{equation}
    \max_{\design}\, \frac{1}{n} \sum_{\klstate \in \klstate_{\text{batch}}} \klq(\klstate, \klpolicy(\klstate, \design), \design)\,,
    \label{Eq::objective}
\end{equation}
with $\klstate_{\text{batch}} = (\klstate_0^1, \klstate_0^2, \cdots, \klstate_0^n)$ containing $n$ randomly chosen start states. 
This objective function can be optimized with classical global optimization methods such as Particle Swarm Optimization (PSO)~\cite{bonyadi2017particle, 494215} or Covariance Matrix Adaptation - Evolution Strategy (CMA-ES) \cite{hansen1996adapting}.

\subsection{Design Generalization and Specialization of Actor and Critic}

A naive solution to input the design variable into the actor and critic network would be to append the design vector to the state and train a single set of networks using the experience of all designs. 
A more promising approach is to have two sets of networks: One \textit{population} (pop.) actor and critic network which is trained on the training experience from all designs, and \textit{individual} (ind.) networks which are initialized with the \textit{population} network but use primarily training experience from the current design (the individual). 
In practice, we found it helpful to allocate 10\% of the training batch for samples from the \textit{population} replay buffer when training the \textit{individual} networks. 
Essentially, this approach allows the \textit{individual} networks $\klq_{\kllocal}$ and $\klpolicy_{\kllocal}$ to specialize in a fast manner to the current design and its nuances to quickly achieve maximum performance. 
In parallel, we are training the \textit{population} networks $\klq_{\klglobal}$ and $\klpolicy_{\klglobal}$ with experience from all designs seen so far by selecting samples 
from the \textit{population} replay buffer $\klreplayglobal$. 
These \textit{population} networks are then able to better generalize across different designs and provide initial weights for the \textit{individual} networks. 
Hence, policies do not have to be learned from scratch for each new prototype. 
Instead, previously collected training data is used so that different designs can inform each other and make efficient use of all the experiences collected thus far.

\subsection{Exploration and Exploitation of Designs}

We alternate between design exploration and exploitation to increase the diversity of explored designs, improve generalization capabilities of the critic and avoid an early convergence to regions of the design space. 
Therefore, every time we find an optimal design during the design optimization process with the objective function (Eq. \ref{Eq::objective}) and conclude the subsequent reinforcement learning process, we next choose one design using the exploration strategy. 
To this end, we implemented two different approaches: sampling new designs 1) randomly, and 2) using Novelty search~\citep{lehman2008exploiting}. 
We found that using random sampling as exploration strategy outperformed novelty search (see appendix). 

\begin{algorithm}[t]
\algsetup{linenosize=\small}
\scriptsize
\begin{algorithmic}
    \STATE Initialize replay buffers: $\klreplayglobal$, $\klreplaylocal$ and $\klreplaystart$ 
    \STATE Initialize first design $\design$
    \FOR{$i \in (1, 2, \cdots, M)$}
        \STATE $\klpolicy_{\kllocal} = \klpolicy_{\klglobal}$
        \STATE $\klq_{\kllocal} = \klq_{\klglobal}$
        \STATE Initialize and empty $\klreplaylocal$
        
        \WHILE{not finished optimizing local policy}
            \STATE Collect training experience $(\klstate_0, \klaction_0, \klreward_1, \klstate_1, \cdots, \klstate_{\text{T}}, \klreward_{\text{T}})$ for current design $\design$ with policy network $\klpolicy_{\kllocal}$
            \STATE Add quadruples $(\klstate_i, \klaction_i, \klreward_{i+1}, \klstate_{i+1})$ to $\klreplaylocal$
            \STATE Add quintuples $(\klstate_i, \klaction_i, \klreward_{i+1}, \klstate_{i+1}, \design)$ to $\klreplayglobal$
            \STATE Add start state $\klstate_0$ to $\klreplaystart$
            
            \STATE Train networks $\klpolicy_{\kllocal}$ and $\klq_{\kllocal}$ with random batches from $\klreplaylocal$
            \STATE Train networks $\klpolicy_{\klglobal}$ and $\klq_{\klglobal}$ with random batches from $\klreplayglobal$
            
        \ENDWHILE
        \IF{$i$ is even}
            \STATE Sample batch of start states $\klstate_{\text{batch}} = (\klstate_0^1, \klstate_0^2, \cdots, \klstate_0^n)$ from $\klreplaystart$
            \STATE \textit{Exploitation}: Compute optimal design $\design$ with objective function $\max_{\design} \frac{1}{n}\sum_{\klstate \in \klstate_{\text{batch}}} \klq_{\klglobal}(\klstate, \klpolicy_{\klglobal}(\klstate, \design), \design)$ 
        \ELSE
            \STATE \textit{Exploration}: Sample design $\design$ with exploration strategy
        \ENDIF
    \ENDFOR
\end{algorithmic}
\caption{\klalgname}
\label{Alg::algorithm}
\end{algorithm}

\subsection{Fast Evolution through Actor-Critic Reinforcement Learning}
The proposed algorithm, \klalgname, is presented in Algorithm \ref{Alg::algorithm}. 
We will now discuss the specifics of the used reinforcement learning algorithm and global optimization method. 
However, it is worth noting that our methodology is agnostic to the specific algorithms used for design and behaviour optimization.  

\paragraph{Reinforcement Learning Algorithm}
While in principal every reinforcement learning method can be employed to train the Q and policy functions necessary to optimize the designs, we use a deep reinforcement learning method due to the continuous state and action domains of our tasks. 
Specifically, we employed the Soft-Actor-Critic (SAC) algorithm \cite{haarnoja2018soft}, a state-of-the-art deep reinforcement learning method based on the actor-critic architecture. 
All neural networks had three hidden layers with a layer size of 200. 
Per episode we train the \textit{individual} networks $\klpolicy_\kllocal$ and $\klq_\kllocal$ 1000 times while the \textit{population} networks $\klpolicy_\klglobal$ and $\klq_\klglobal$ are trained 250 times. 
The motivation was to assign more processing power to the \textit{individual} networks to adapt quickly to a design and specialize. 
A batch size of 256 was used for each training updated. 

\paragraph{Optimization Algorithm}
\label{Sub::OptAlg}
To optimize the objective function given in \eq{Eq::objective}, we used the global optimization method Particle Swarm Optimization (PSO)~\cite{bonyadi2017particle, 494215}. 
We chose PSO primarily because of its ability to search the design space exhaustively using a large number of particles. 
The objective function (\eq{Eq::objective}) was optimized using about 700 particles, each representing a candidate design, and updated over 250 iterations. 
Accordingly, PSO used a total contingent of 175,000 objective function evaluations to find an optimal design. 
To optimize the design using rollouts in simulation, we had to reduce this number to about 1,050 design candidates, i.e. 35 particles updated over 30 iterations. 
Although this contingent is only about $0.6\%$ of the size of the Q-function contingent, it takes about two times longer to evaluate this number of designs in simulation. 
For example, on a system with an Intel Xeon CPU E5-2630 v4 CPU equipped with an NVIDIA Quadro P6000, the design optimization via simulation takes approximately 30 minutes while the optimization routine using the critic requires only 15 minutes. 
To put this into perspective, the reinforcement learning process on a single design requires approximately 60 minutes for 100 episodes.


\section{Experimental Evaluation}
\label{sec:result}

We now experimentally evaluate our proposed approach, with the aim of answering the following questions:
1) Can we obtain with our algorithm comparable task performance as optimizing the design by performing extensive trials, by instead relying on the learned model?
2) If so, how much can our approach reduce the number of trials?
3) Can our approach help us to get insight into the design space that we are trying to optimize for a specific task?

Code for reproducing the experiments, videos, and additional material is available online at\\ \url{https://sites.google.com/view/drl-coadaptation}. 

\subsection{Experimental Setting}

\begin{figure}
    \centering
    \begin{subfigure}[b]{0.24\linewidth}
        \includegraphics[width=\textwidth]{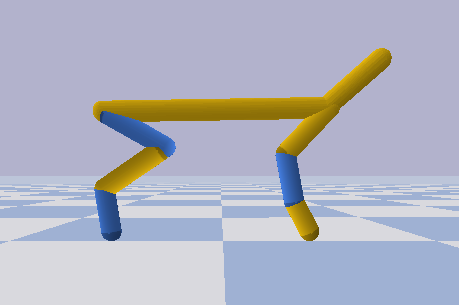}
        \caption{Half Cheetah}
        \label{Fig::Tasks::HC}
    \end{subfigure}
    \hfill
    \begin{subfigure}[b]{0.24\linewidth}
        \includegraphics[width=\textwidth]{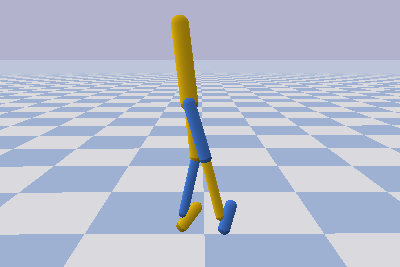}
        \caption{Walker}
        \label{Fig::Tasks::Walker}
    \end{subfigure}
    \hfill
    \begin{subfigure}[b]{0.24\linewidth}
        \includegraphics[width=\textwidth]{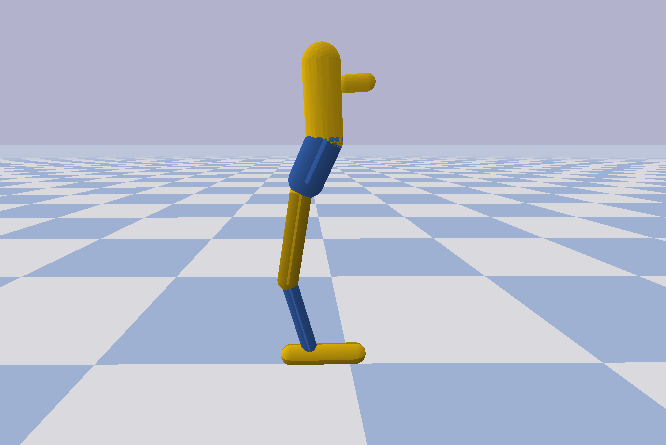}
        \caption{Hopper}
        \label{Fig::Tasks::Hopper}
    \end{subfigure}
    \hfill
    \begin{subfigure}[b]{0.24\linewidth}
        \centering
        \includegraphics[width=\textwidth]{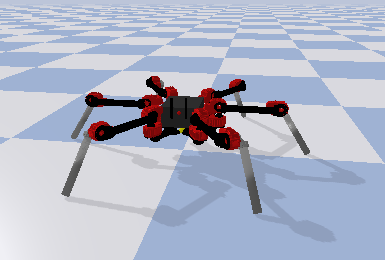}
        \caption{Daisy Hexapod}
        \label{Fig::Tasks::Daisy}
    \end{subfigure}
    \caption{The four simulated robots used in our experiments.}
    \label{Fig::Tasks}
\end{figure}

To evaluate our algorithm, we considered the four control tasks simulated using PyBullet~\cite{coumans2019} shown in \fig{Fig::Tasks}.
The design of agents for each task is described as a continuous design vector $\design \in \mathbb{R}^d$.
The initial five designs for each task were pre-selected with the original design and four randomly chosen designs which were consistent over all experiments. 
All experiments were repeated five times. 
For the standard PyBullet tasks (\Cref{Fig::Tasks::HC,Fig::Tasks::Walker,Fig::Tasks::Hopper}) we executed 300 episodes for the initial five designs and 100 episodes thereafter.
The latter was increased to 200 episodes for the more complex Daisy Hexapod task (\fig{Fig::Tasks::Daisy}) \cite{hebi}. 
We will give a short description of the simulated locomotion tasks and state for each task the number of states, actions and design parameters as a vector $(\klstate, \klaction, \design)$. 
A detailed descriptions of the tasks can be found in the appendix. 
\textbf{Half-Cheetah (17, 6, 6)} and \textbf{Walker (17, 6, 6)} are agents with two legs tasked to learn to run forward. 
Each agent has six leg segments to be optimized independently for their length. 
The \textbf{Hopper (13, 4, 5)} agent has a single leg with four leg segments as well as a nose-like feature and has to learn to move forward as well. 
All three agents are restricted to movements in a 2D plane. 
The \textbf{Daisy Hexapod (43, 18, 9)} simulates an hexapod and is able to move in all three dimensions. 
Its goal is to learn to move forward without changing its orientation. 
The lengths of the leg-segments are mirrored between the left and right side of the robot, with three leg-segments per leg.

\subsection{Co-adaptation Performance}
\label{Sec::Exp::General_Performance}

\begin{figure}[t]
    \centering
    \includegraphics[width=\textwidth]{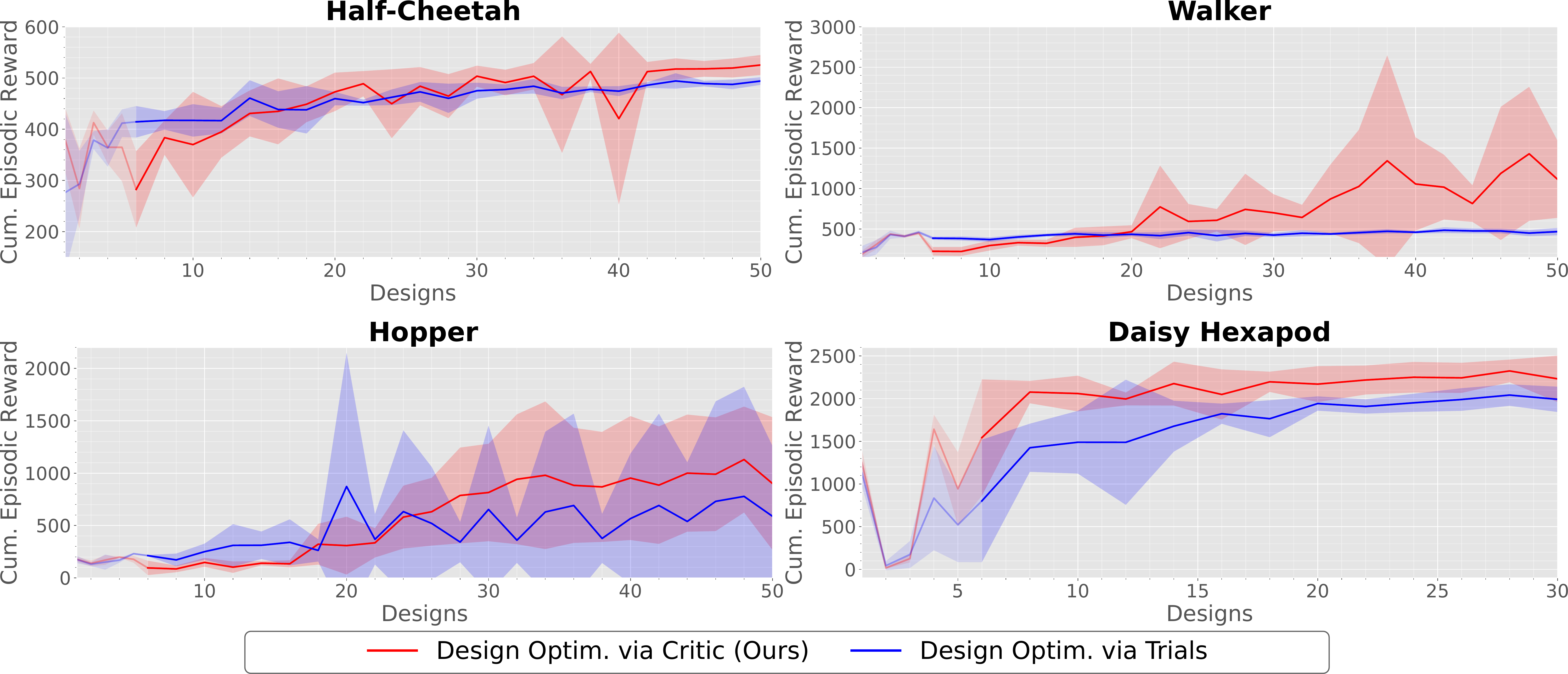}
    \caption{Comparison between our proposed approach (red) and using trials to evaluate the optimality of candidate designs (blue). 
    The plots each show the mean and standard deviation of the highest reward achieved over five experiments for optimal designs $\design_{\text{Opt}}$. 
    We can see that the proposed method (\fig{Fig::concepts::proposed}) has a comparable or even better performance than optimizing designs via executions in simulation (\fig{Fig::concepts::normal}). 
    }
    \label{Fig::Sim_vs_Q}
\end{figure}

\begin{figure}[t]
    \centering
    \begin{subfigure}[b]{0.49\textwidth}
        \includegraphics[width=\textwidth]{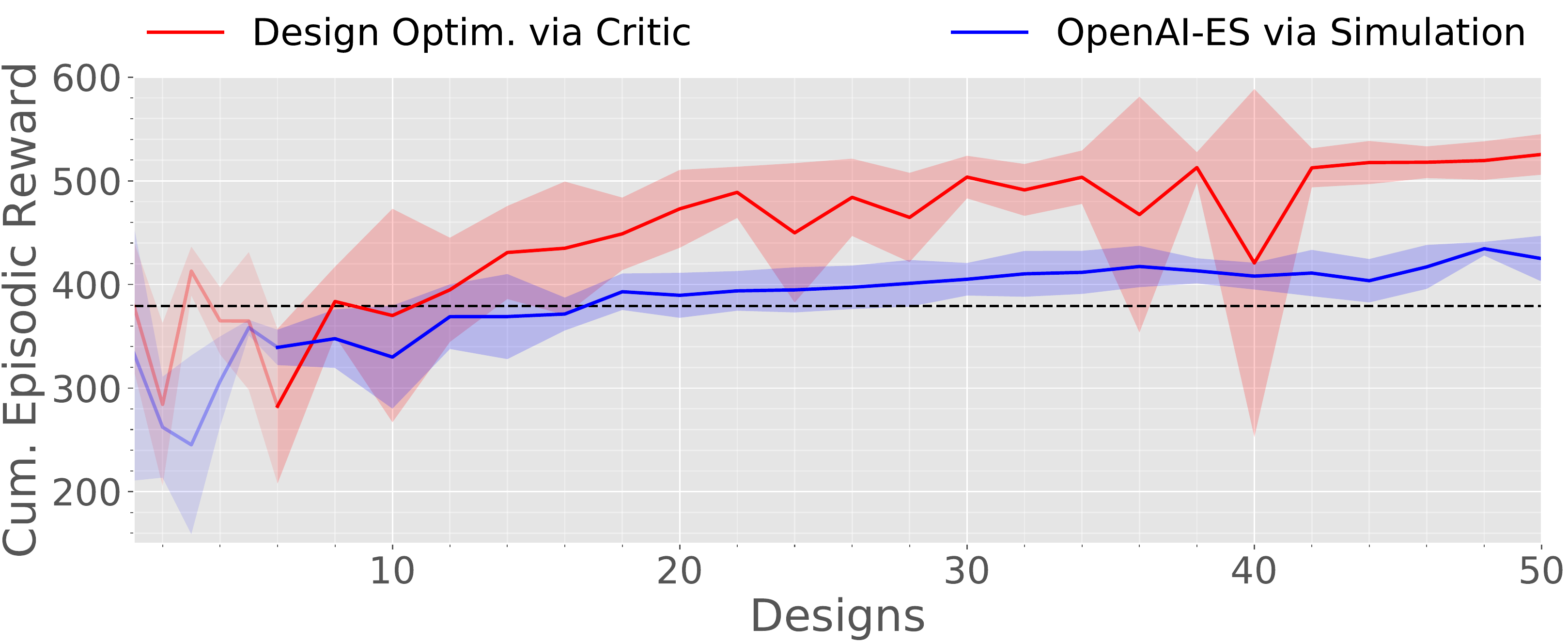}
        \label{Fig::Opt_and_Random::OpenAIes}
    \end{subfigure}
    \hfill
    \begin{subfigure}[b]{0.49\textwidth}
        \includegraphics[width=\textwidth]{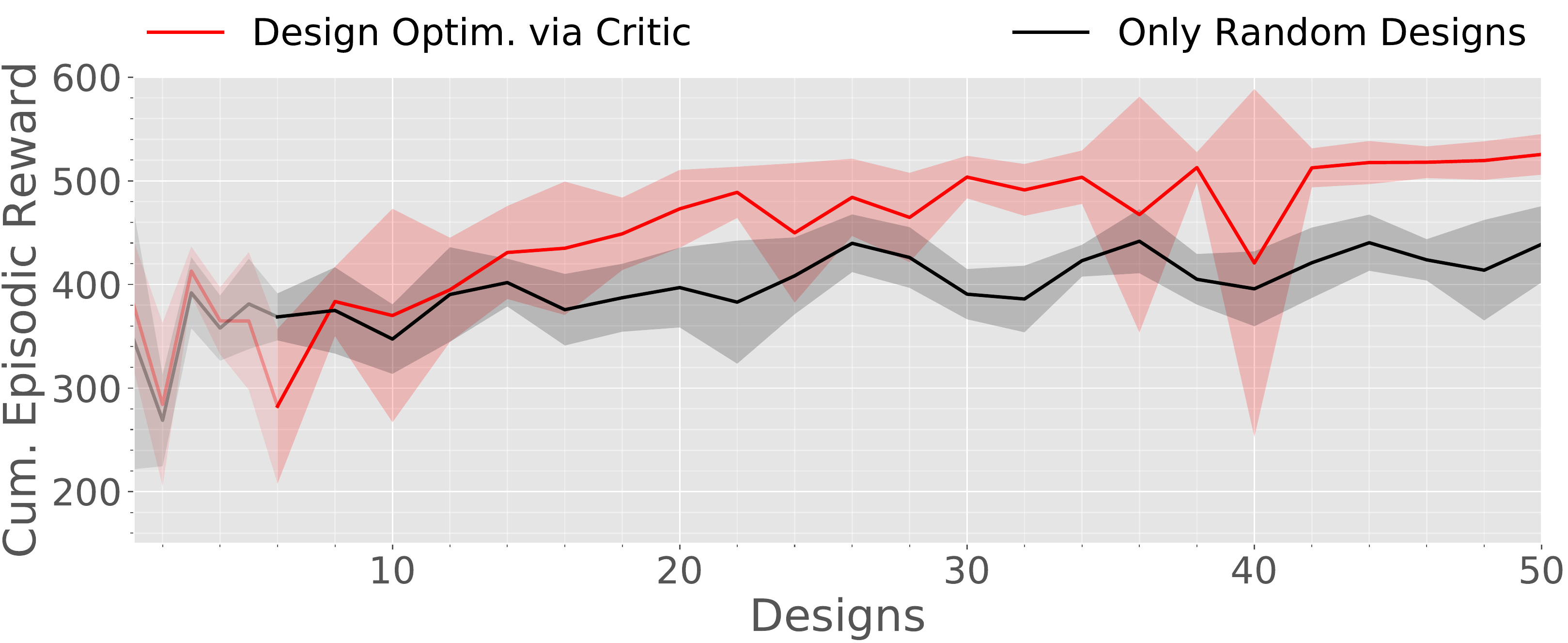}
        \label{Fig::Opt_and_Random::Random}
    \end{subfigure}

    \caption{ 
    (\textit{Left}) Comparison on the Half-Cheetah against the proposed method and OpenAI-ES~\cite{salimans2017evolution} with a population size of 256 as used in \cite{ha2018reinforcement}. While our method uses the proposed objective function, OpenAI-ES uses the simulator for evaluating design candidates. The dotted line shows the average reward achieved on the original design of Half-Cheetah. 
    (\textit{Right}) Comparison of the proposed method and sampling \textbf{only} random designs instead of optimizing the objective function. 
    The plots show the mean and standard deviation of the highest reward achieved over five experiments. The proposed approach outperforms the random baseline. 
    }
    \label{Fig::Opt_and_Random}
\end{figure}

We compared the proposed framework, using actor-critic networks for design evaluation, and the classical approach, optimizing the design through candidate evaluations in simulation, on all four locomotion tasks (\fig{Fig::Sim_vs_Q}). 
We can see that, especially in the Half-Cheetah task, using actor-critic networks might perform worse over the first few designs but quickly reaches a comparable performance and even surpasses the baseline. 
It is hypothesized that the better performance in later episodes is due to the ability of the critic to interpolate between designs while the evaluations of designs in simulation suffers from noise during execution. 
Interestingly, using simulations to optimize the design does not seem to lead to much improvement in the case of the Walker task. 
This could be due to the randomized start state, which often leads to the agent being in an initial state of falling backwards or forwards, which would have an immediate effect on the episodic reward. 
Additionally, we compared the proposed method using the introduced objective function for evaluating design candidates against the method used for design optimization in \cite{ha2018reinforcement}. 
\fig{Fig::Opt_and_Random} shows that the evolution strategy OpenAI-ES~\cite{salimans2017evolution}, using the simulator to evaluate design candidates with a population size of 256, is outperformed by our proposed method. %
Moreover, we verified that for all experiments, designs selected randomly, with a uniform distribution, performed worse than designs selected through optimization (see \fig{Fig::Opt_and_Random}).

\begin{figure}[t]
    \centering
    \includegraphics[width=\textwidth]{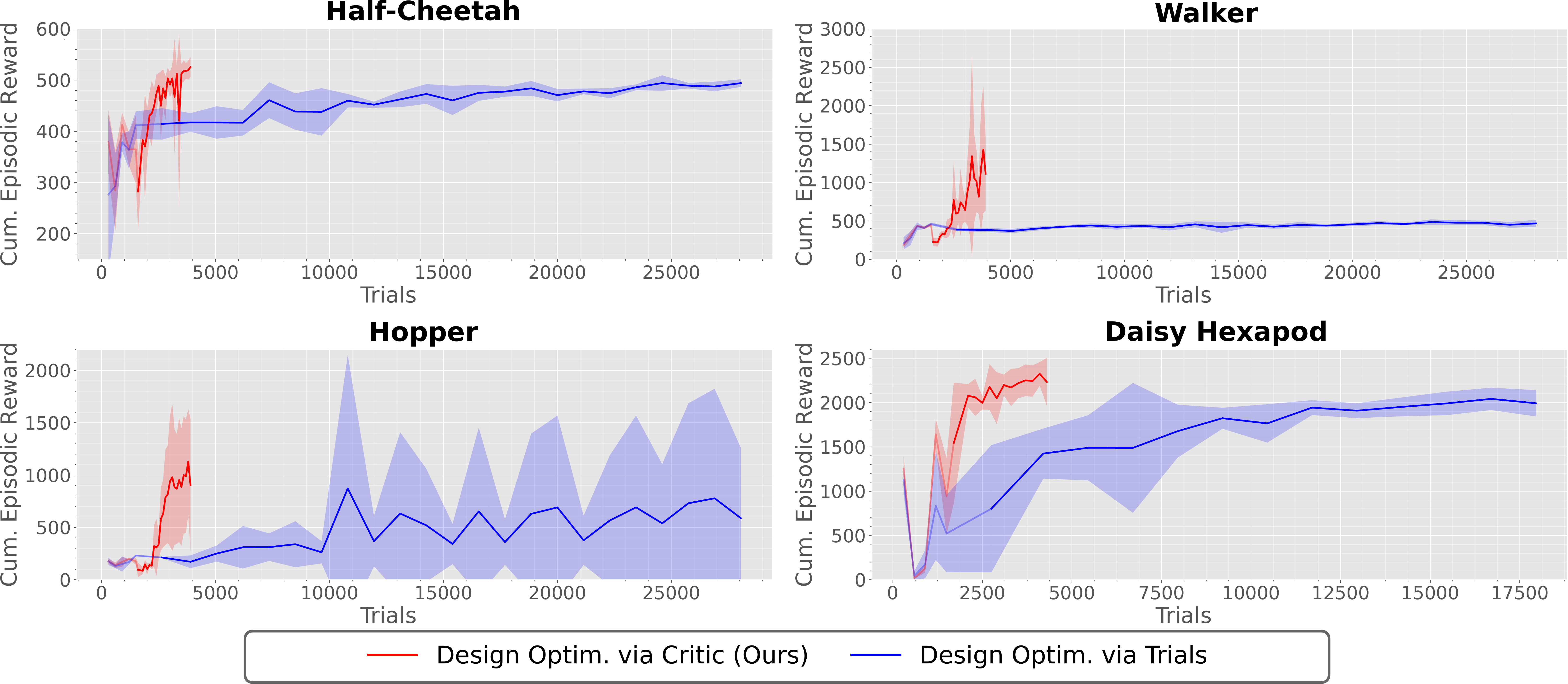}    
    \caption{Comparison between the our proposed approach of using the actor and critic to optimize the design parameters (red) and using trials/simulations to evaluate the optimality of candidate designs (blue). 
    The plots show the mean and standard deviation of the highest reward achieved over five experiments. 
    The x-axis shows the number of episodes executed in simulation. 
    We can see that removing the need to simulate design candidates during the design optimization process leads to a comparable performance in a much shorter time frame.  
    }
    \label{Fig::Nmbr_Simulations}
\end{figure}

\paragraph{Simulation Efficiency}
\label{Sec::Exp::Simulation_Efficiency}
To evaluate the suitability of the proposed method for deployment in the real world, we compared the methods based on the number of simulations required. 
As we can see in \fig{Fig::Nmbr_Simulations}, the actor-critic approach quickly reaches a high performance quickly with a low number of simulations. 
As explained above, this is due to the design optimization via simulation requiring 1,050 simulations to find an optimal design while the proposed method requires none. 

\begin{figure}[t]
    \centering
    \includegraphics[width=\textwidth]{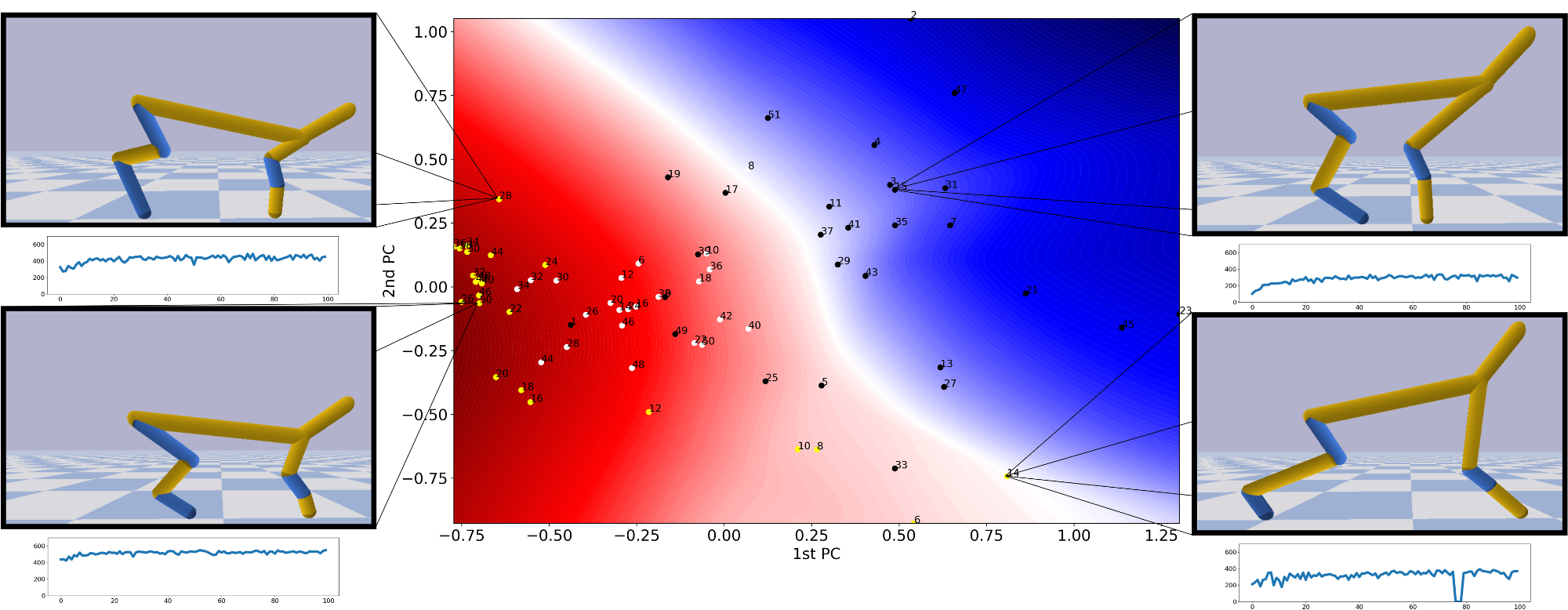}
    \caption{First two principal components of the six dimensional design space of Half-Cheetah as computed with PCA. 
    Colours indicate the Q-value given by the critic on a batch of 256 start states after 50 evaluated designs, with red indicating regions of higher expected reward, and blue the regions of low expected reward. 
    The designs chosen by our approach are depicted as yellow dots, the white dots are the designs selected when optimizing via simulation, and the black shows randomly selected design. 
    Numbers indicate the order in which the designs were chosen for reinforcement learning.}
    \label{Fig::Latent_Space}
\end{figure}

\paragraph{Visualization of Reward Landscapes for Designs}
\label{Sec::Exp::Visualization}
A major advantage of the proposed method is the possibility to visualize the expected reward for designs. 
Instead of selecting a number of designs to evaluate, which would take a significant effort in the real world as well as computationally, we are able to query the introduced objective function (\eq{Eq::objective}) in a fast manner. 
This allows us to visually inspect the reward landscape of designs and to gai ninsight at what makes designs perform better or worse. 
In \fig{Fig::Latent_Space}, the first two principal components were computed based on the designs selected for learning in the Half-Cheetah task. 
We can see, for example, that a shorter second segment of the back leg and as well as a shorter first segment of the front leg seems to be desirable.

\section{Conclusion}
\label{sec:conclusion}
In this paper, we study the problem of data-efficiently co-adapting morphologies and behaviors of robots.
Our contribution is a novel algorithm, based on recent advances in deep reinforcement learning, which can better exploit previous trials to estimate the performance of morphologies and behaviors before testing them.
As a result, our approach can drastically reduce the number of morphology designs tested (and their eventual manufacturing time/cost).
Experimental results on 4 simulated robots show strong performance and a drastically reduced number of design prototypes, with one robot requiring merely 50 designs compared to the 24,177 of the baseline -- that is about 3 orders of magnitudes less data.
The unparalleled data-efficiency of our approach opens exciting venues towards the use in the real world of robots that can co-adapt both their morphologies and their behaviors to more efficiently learning to perform the desired tasks with minimal expert knowledge. 
In future work, we aim to demonstrate the capabilities of this algorithm on a robot in the real world.



\clearpage
\acknowledgments{
We thank Akshara Rai for the valuable discussions during the early stages of this research, as well as for testing the early implementations of the Daisy hexapod simulation thoroughly. 
Furthermore, we thank Ge Yang for his support to run additional simulations when they were needed.  
Finally, we thank the anonymous reviewers for their helpful comments.  
}

\bibliography{content/papers}  

\clearpage

\section{Appendix}

\subsection{Simulation Environments}
This section states a short description of each task simulated in PyBullet \cite{coumans2019}:
\paragraph{Half-Cheetah (17, 6, 6)}
The half cheetah task has an 17 dimensional state space consisting of joint positions, joint velocities,  horizontal speed, angular velocity, 
 vertical speed and relative height. 
 Actions have six dimensions and are accelerations of joints. 
The original reward function used in PyBullet was adapted to be design independent and is given by $r(\klstate) = \max(\frac{\Delta x}{10}, 0)$ where $\Delta x$ is the horizontal speed to encourage forward motion. 
The continuous design vector is a scaling factor of the original leg lengths of Half-Cheetah: 
$(\design_1 \cdot 0.29, \design_2 \cdot 0.3, \design_3 \cdot 0.188, \design_4 \cdot 0.29, \design_5 \cdot 0.3, \design_6 \cdot 0.188)$. 
The dimensions of the design vector are in the interval $\design_i \in [0.8, 2.0]$. 

\paragraph{Walker (17, 6, 6)}
Similar to the Half-Cheetah task, the state space of the Walker task is given by joint positions, joint velocities,  horizontal speed, angular velocity, 
 vertical speed and relative height and has 16 dimensions. 
The two legs of Walker are controlled through acceleration with a six dimensional action. 
Again, the original reward was adapted to be design agnostic. 
The term encouraging maximum height of the torso of walker was replaced by two terms favouring vertical orientation $y_\text{rot}$ of the torso and reaching a minimal height $h_{\text{torso}}$ of $0.8$. 
The full reward function is given by $\klreward(\klstate) = \frac{1}{10} \left( (h_{\text{torso}} > 0.8) \cdot \left( \max(\Delta x, 0) + 1 \right) - \parallel y_\text{rot}\parallel_2 0.1\right)$. 
The design vector is a scaling factor of the leg and foot lengths of the Walker agent: 
$(\design_1 \cdot 0.45, \design_2 \cdot 0.5, \design_3 \cdot 0.2, \design_4 \cdot 0.45, \design_5 \cdot 0.5, \design_6 \cdot 0.2)$. 
Each design dimension lies in the interval $\design_i \in [0.5, 1.5]$. 

\paragraph{Hopper (13, 4, 5)}
In the planar Hopper task a one-legged agent has to learn jumping motions in order to move forward. The state space of this task has thirteen dimensions and four dimensions in the action space. 
We use the same reward function as for the Walker task with $\klreward(\klstate) = \frac{1}{10} \left( (h_{\text{torso}} > 0.8) \cdot \left( \max(\Delta x, 0) + 1 \right) - \parallel y_\text{rot}\parallel_2 0.1\right)$. 
In addition to the length of the four movable leg segments, the length of the nose-like feature of walker is an additional design parameter, here $\design_1$. 
The full design vector is given by $\design = (\design_1 \cdot 0.7, \design_2 \cdot 0.15, \design_3 \cdot 0.33, \design_4 \cdot 0.32, \design_5 \cdot 0.25)$ with $\design_{2:5}$ being the length of each movable segment from pelvis to foot. 
The design parameters were bounded with $\design_1 \in [0.5, 4.0]$ for the length of the nose and $\design_{2:5}\in[0.5, 2.0]$ for all leg lengths. 

\paragraph{Daisy Hexapod (43, 18, 9)}
For a preliminary study and to evaluate whether the proposed method is suitable for real world applications, a simulation of the six-legged Daisy robot by HEBI Robotics \cite{hebi} was created in PyBullet. 
Each leg of the robot has three motors and hence the action space has 18 dimensions. 
The state space has 43 dimensions and consists of joint positions, joint velocities, joint accelerations, the velocity of the robot in x/y/z directions and the orientation of the robot in Euler angles. 
The task of the robot is to learn to walk forward while keeping its orientation and thus the reward function is given by $r(s) = \frac{\max(\Delta y, 0)}{0.066} - 0.25 \cdot \text{diff}(e_\text{original}, e_\text{current})$, with $\Delta y$ being the dislocation along the y-axis, the direction the robot faces at initialization, and $\text{diff}(e_\text{original}, e_\text{current})$ representing the angle between the original and current orientation in quaternions. 
The design vector consists of two parts: leg lengths, and movement range of the motors at the base of the legs. 
All parameters are symmetric between the left and right side of the robot. 
The leg lengths are in $\design_{1:6} \in [0.12, 0.5]$ for the two leg segments of each leg. 
Additionally, we allowed the algorithm to optimize the movement range of the first out of three motors on each leg. 
The base motors are restricted in movement between $(-0.35 + \design_{7:9}, 0.35 + \design_{7:9})$ radians with the design parameters $\design_{7:9}\in [-0.2, 0.2]$.

\subsection{Visualization of Design Space}
\begin{figure}[h]
    \centering
    \begin{subfigure}[b]{0.49\textwidth}
        \includegraphics[height=0.6\textwidth]{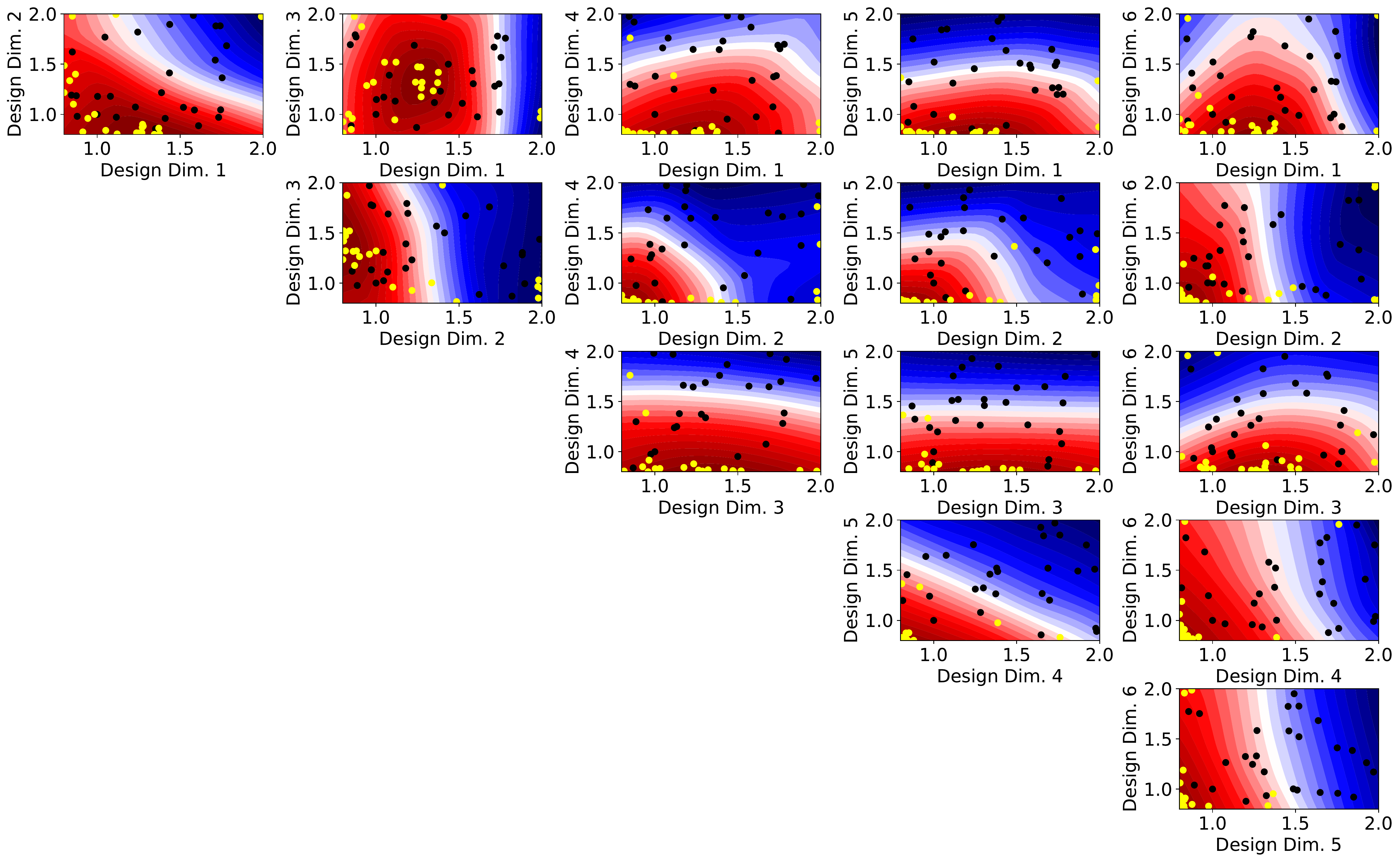}
        \caption{Design Space of Half Cheetah for \newline $\design = (1.27,0.87,1.17,0.84,0.80,0.82)$}
    \end{subfigure}
    ~%
    \begin{subfigure}[b]{0.49\textwidth}
        \includegraphics[height=0.6\textwidth]{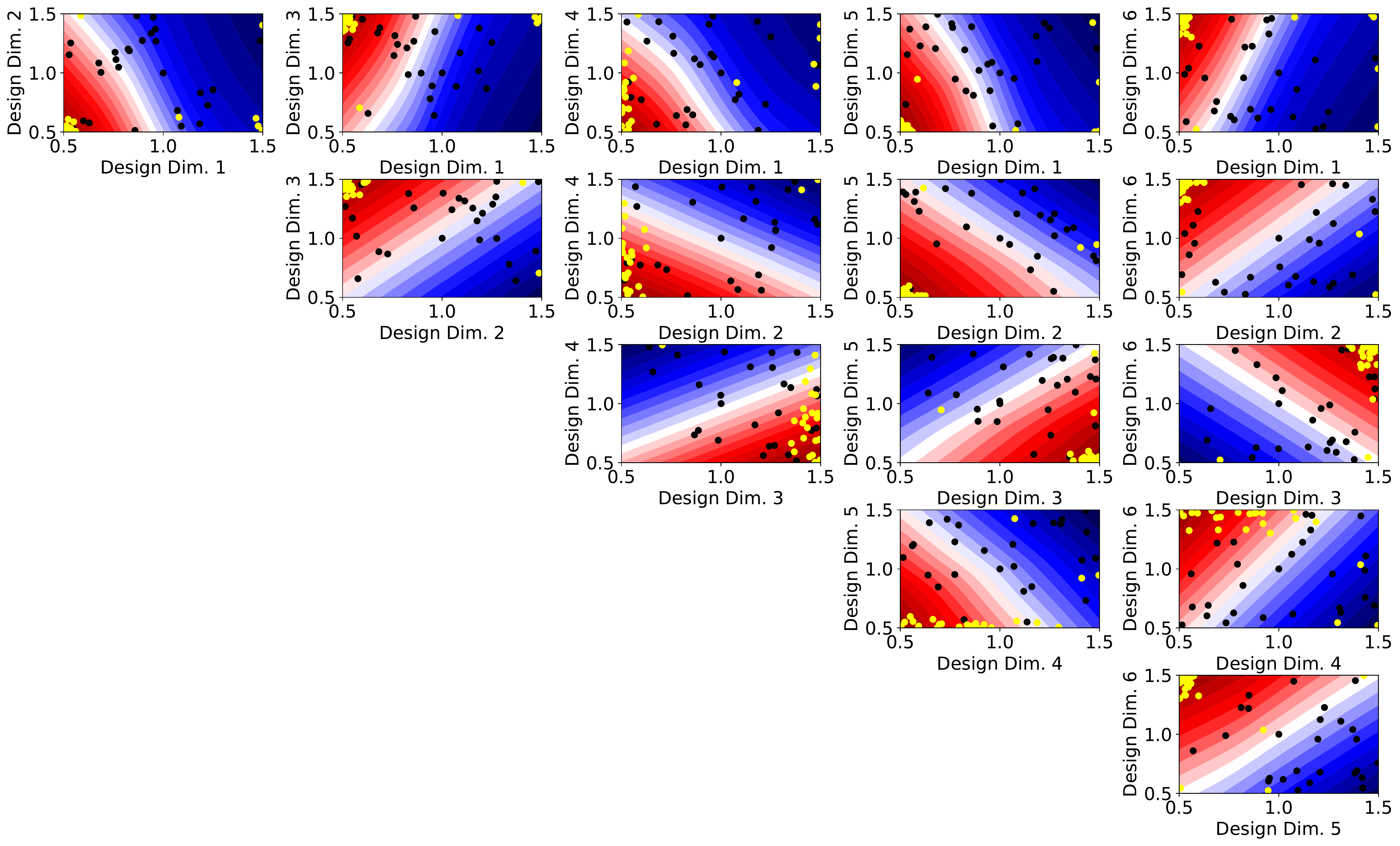}
        \caption{Design space of Walker for \newline $\design = (0.50,0.52,1.35,0.66,0.57,1.49)$}
    \end{subfigure}
    \begin{subfigure}[b]{0.49\textwidth}
        \includegraphics[height=0.6\textwidth]{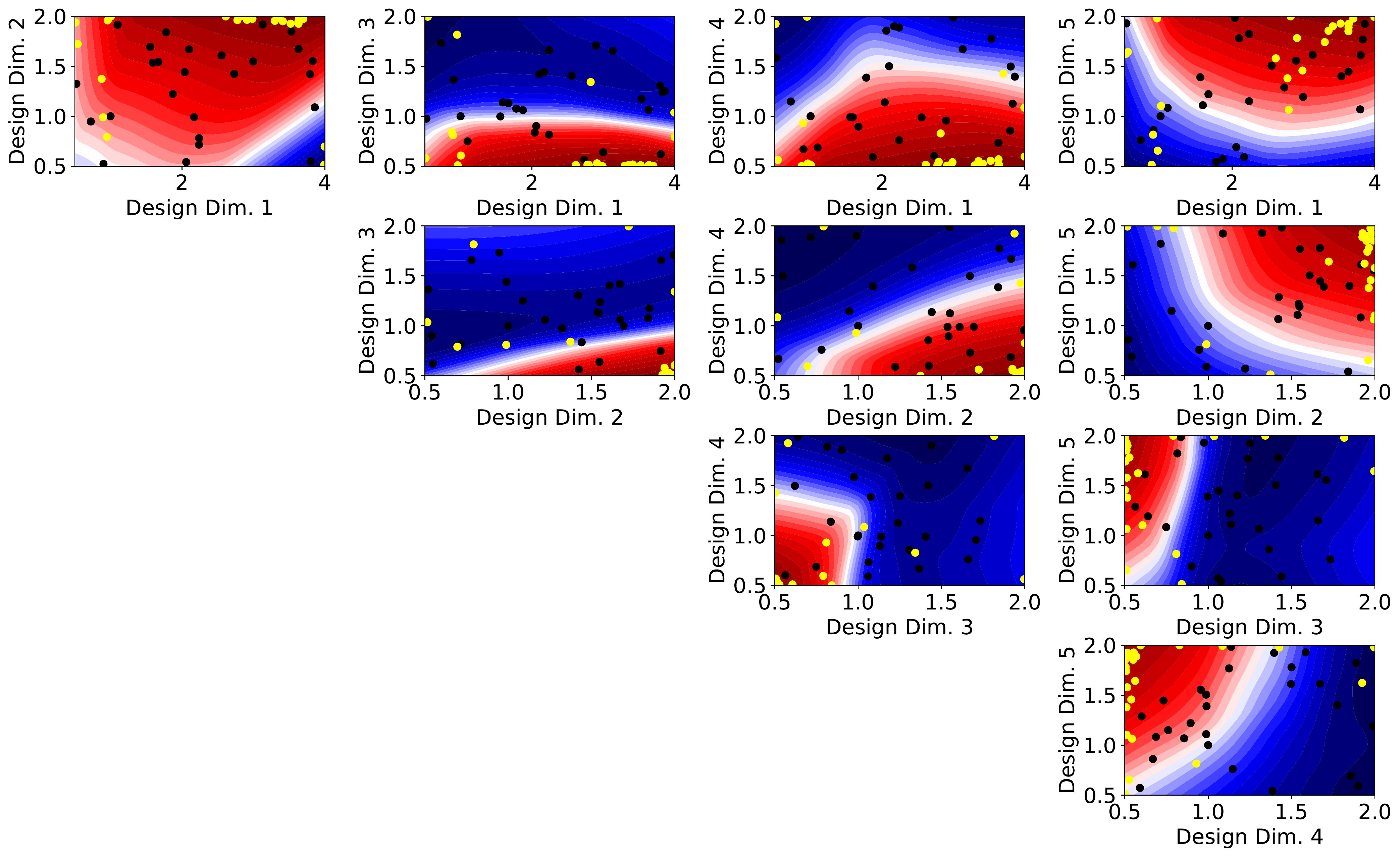}
        \caption{Design Space of Hopper for \\ $\design = (3.41,1.95,0.52,0.52,1.90)$}
    \end{subfigure}
    \caption{The visualized cost landscape of the design spaces. A batch of 256 start states was used.}
    \label{Appendix::Fig::design_spaces}
\end{figure}

\begin{figure}[h!]
    \centering
    \includegraphics[height=0.6\textwidth]{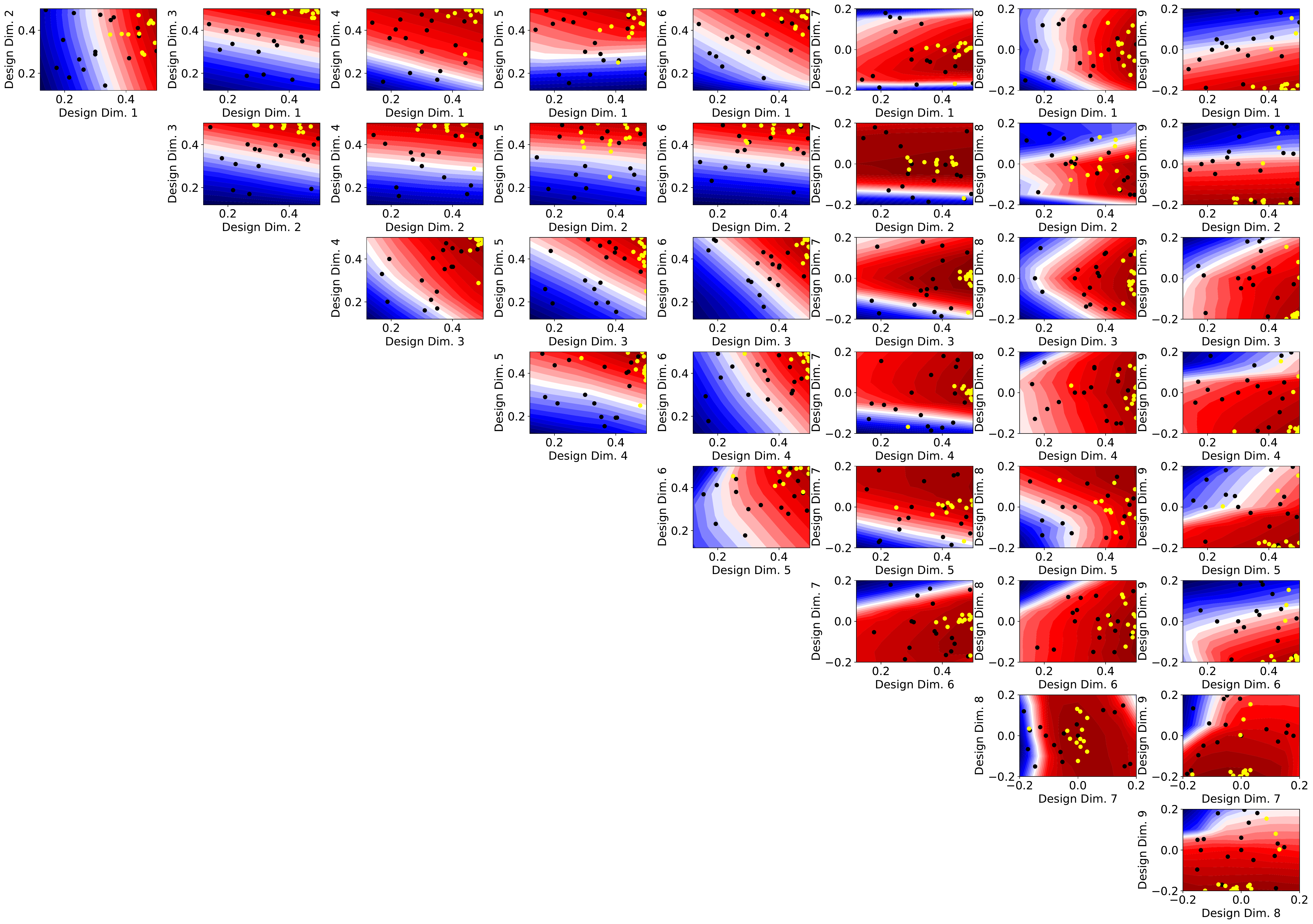}
    \caption{Design space of the Daisy Hexapod for $\design = (1.27,0.87,1.18,0.84,0.80,0.83)$. A batch of 256 start states was used.
    The designs chosen by our approach are depicted as yellow dots, the white dots are the designs selected when optimizing via simulation, and the black shows randomly selected design.}
    \label{Appendix::Fig::Daisy_design_space}
\end{figure}

Because we can query the proposed objective function from eq. \ref{Eq::objective}, we are able to visualize the cost landscape of each task. 
Figure \ref{Appendix::Fig::design_spaces} shows the design spaces of the three standard PyBullet tasks Half-Cheetah, Walker and Hopper after 50 designs evaluated in simulation. 
Each single plot shows the design landscape of two dimensions while the other dimensions were held fix with stated design vectors as well as the location of design chosen by the proposed method (yellow) and designs chosen randomly for exploration (black). 
The cost landscape of the more complex Daisy Hexapod task is shown in figure \ref{Appendix::Fig::Daisy_design_space}. 

\newpage
\subsection{Visualization of the Latent Design Space}
\begin{figure}
    \centering
    \begin{subfigure}[b]{0.49\textwidth}
        \includegraphics[width=\textwidth]{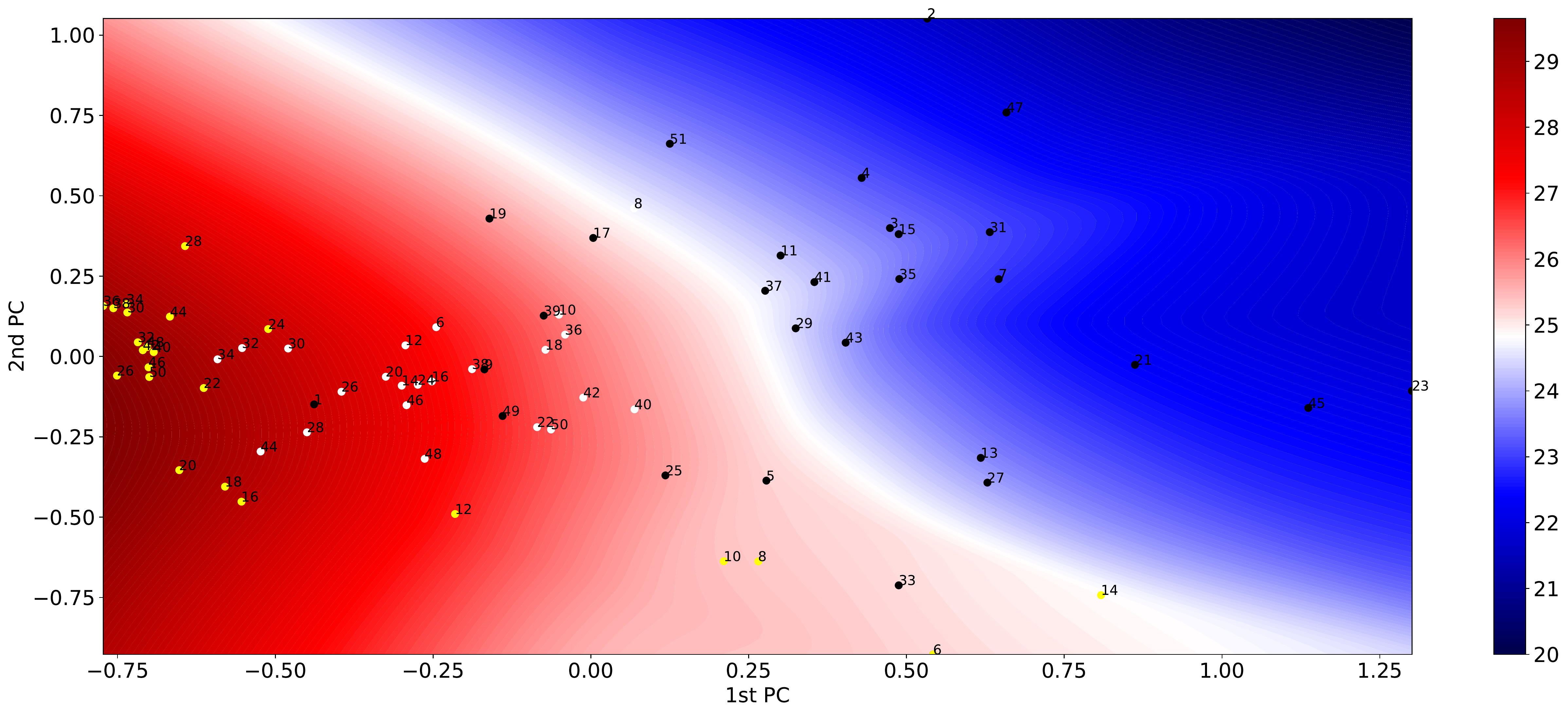}
        \caption{Half-Cheetah}
        \label{Appendix::Fig::latent_design_spaces::HC}
    \end{subfigure}
    ~%
    \begin{subfigure}[b]{0.49\textwidth}
        \includegraphics[width=\textwidth]{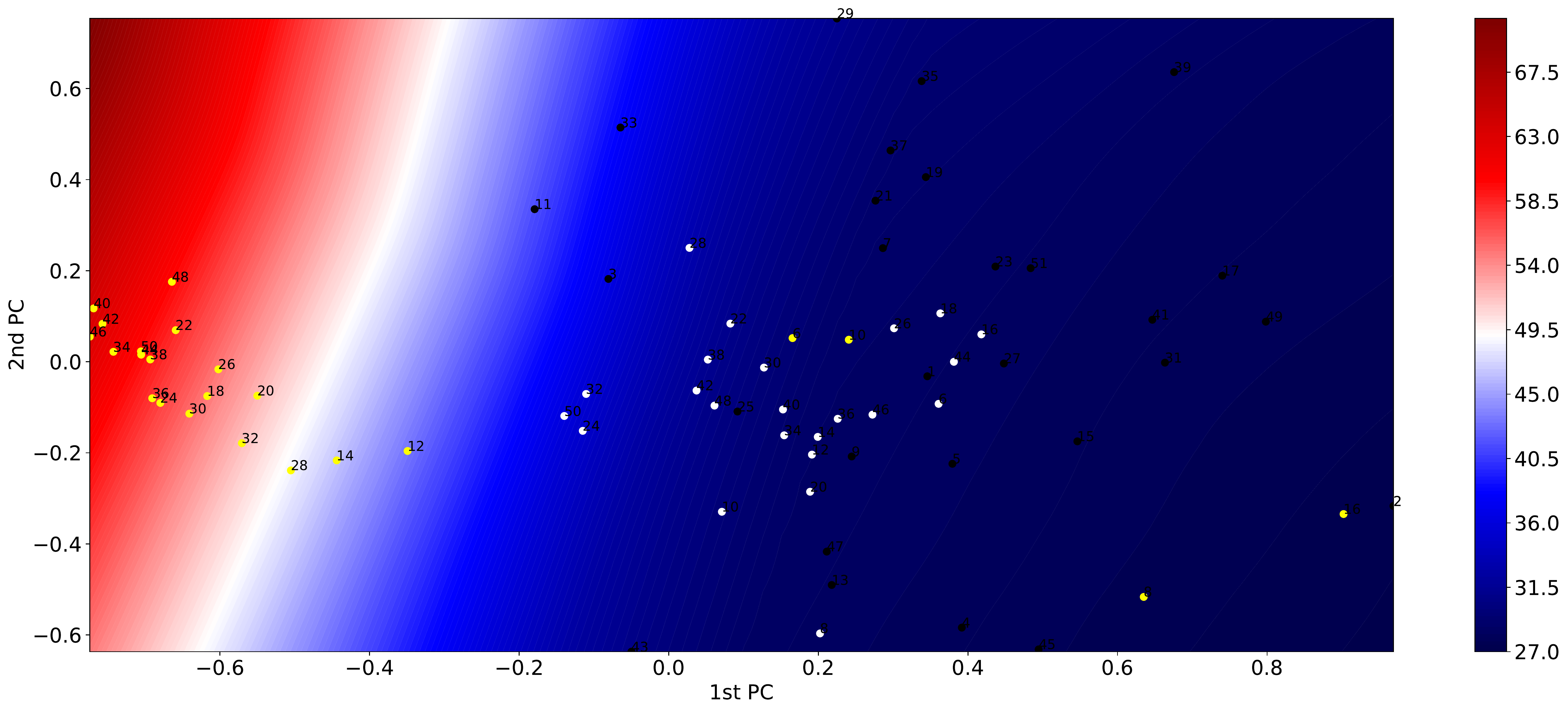}
        \caption{Walker}
    \end{subfigure}
    \begin{subfigure}[b]{0.49\textwidth}
        \includegraphics[width=\textwidth]{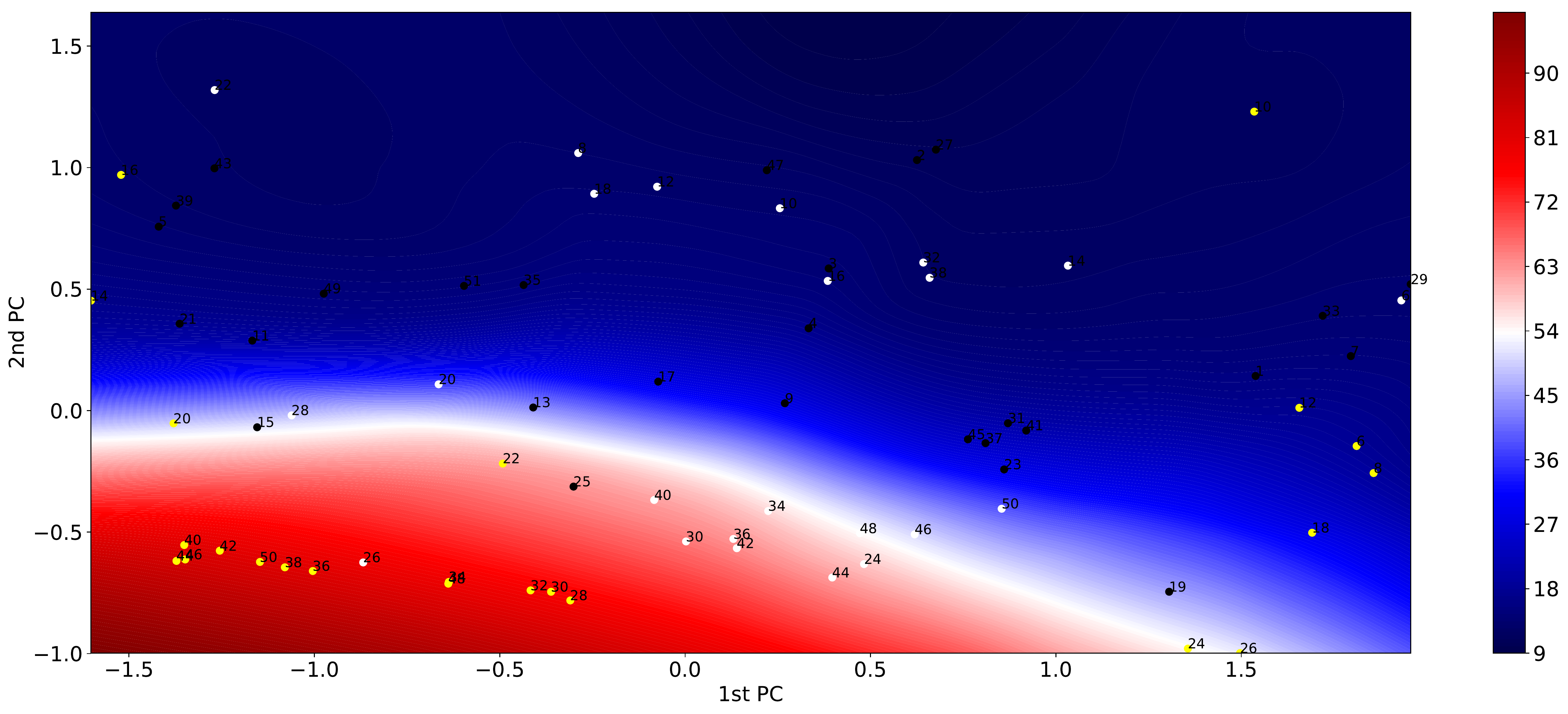}
        \caption{Hopper}
    \end{subfigure}
    ~%
    \begin{subfigure}[b]{0.49\textwidth}
        \includegraphics[width=\textwidth]{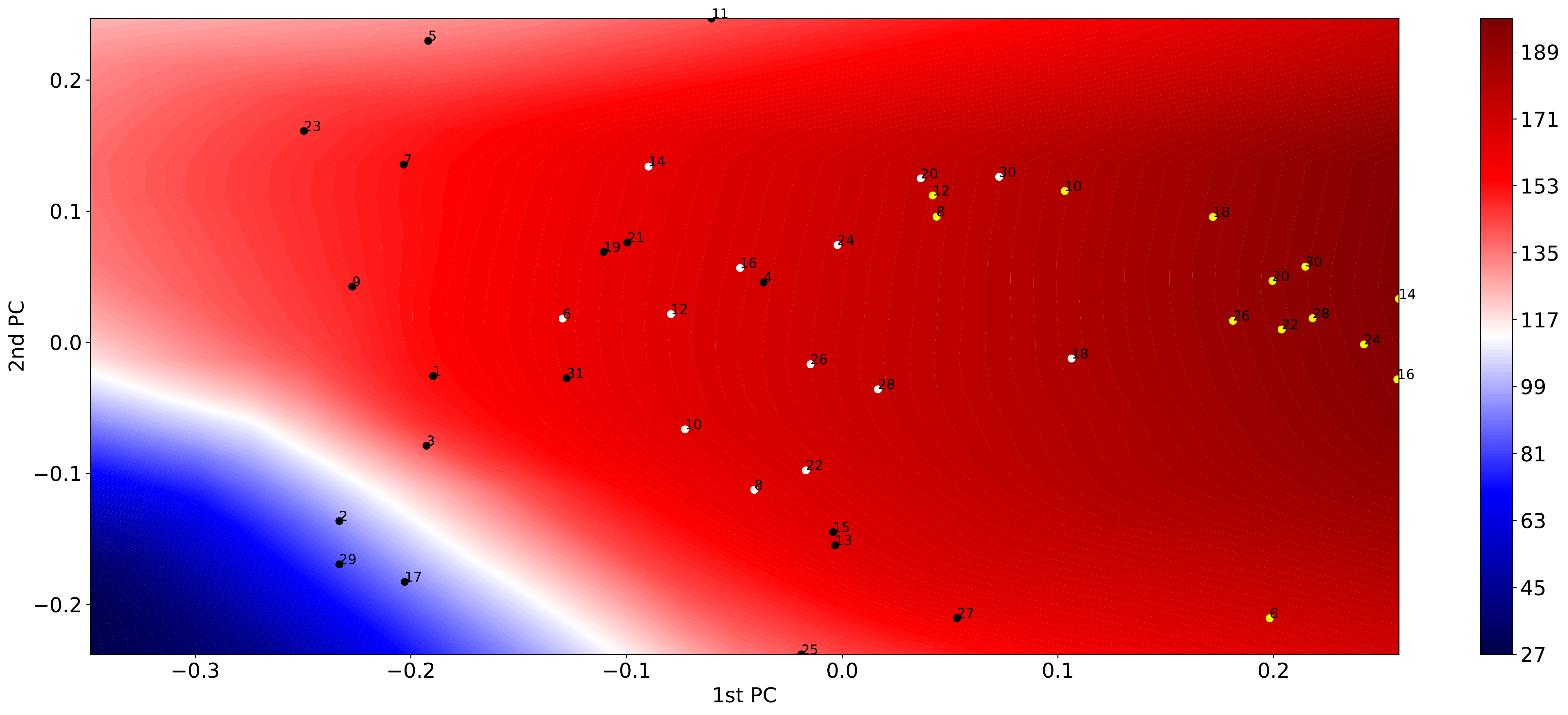}
        \caption{Daisy Hexapod}
    \end{subfigure}
    \caption{First two principal components of each design space as computed with PCA. 
    Colours indicate the Q-value given by the critic on a batch of 256 start states after 50 (30 for the Daisy Hexapod) evaluated designs, with red indicating regions of higher expected reward, and blue the regions of low expected reward. 
    The designs chosen by our approach are depicted as yellow dots, the white dots are the designs selected when optimizing via simulation, and the black shows randomly selected design. 
    Numbers indicate the order in which the designs were chosen for reinforcement learning.}
    \label{Appendix::Fig::latent_design_spaces}
\end{figure}

\begin{figure}
    \centering
    \includegraphics[width=0.98\textwidth]{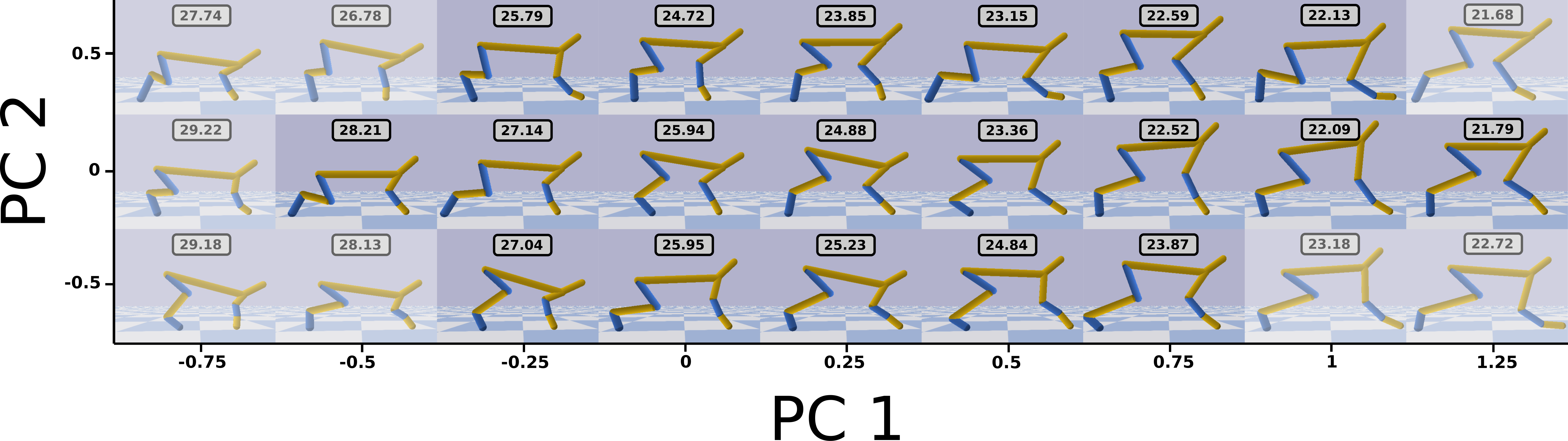}
    \caption{A selection of designs for Half-Cheetah, generated from the principal components in Fig. \ref{Appendix::Fig::latent_design_spaces::HC}. 
    Designs which are outside of the bounds set for the design space are reduced in opacity. 
    Each design is evaluated with the objective function stated in Eq. \ref{Eq::objective}. }
    \label{Appendix::Fig::latent_space_designs}
\end{figure}

For a better understanding of the cost landscape a low dimensional design space was computed with principal component analysis. 
Figure \ref{Appendix::Fig::latent_design_spaces} shows the low-dimensional projection of the design space as well as the designs $\design_{\text{Opt}}$ chosen by the proposed method (ywllow) and randomly selected designs for exploration (black)
In white designs chosen by the optimization via simulation method are shown. 
We can see that the convergence rate of \textit{optimization via simulation} appears to be slower than our method. 
To see what properties of the design lead to a better performance we visualized the design along the two principal components (Fig. \ref{Appendix::Fig::latent_space_designs}). 
We can see that just longer leg do not appear to lead automatically to better performance but shorter front legs and slightly longer back legs do. 

\subsection{Evolution of Walker}
Figure \ref{Appendix::Fig::walker_designs} shows the evolution of designs with the proposed objective function. 
We can see that the start states are random and lead to different poses of Walker, sometimes falling for- or backwards. 
It can be seen that while shorter legs seem desirable, the larger the foot length the better the performance. 
{

\begin{figure}[tbp]
    \centering
    \includegraphics[width=0.98\textwidth]{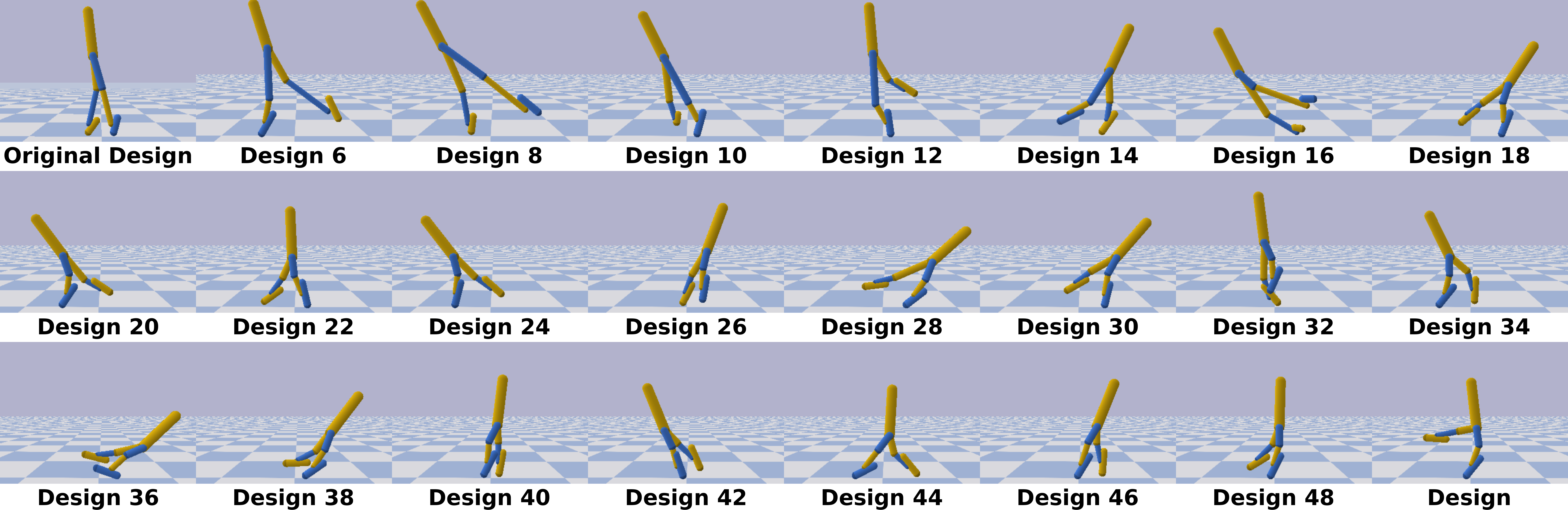}
    \caption{Designs $\design_{\text{Opt}}$ selected by the proposed method for the Half-Cheetah task.}
    \label{Appendix::Fig::walker_designs}
\end{figure}
}

\begin{figure}
    \centering
    \includegraphics[width=0.6\textwidth]{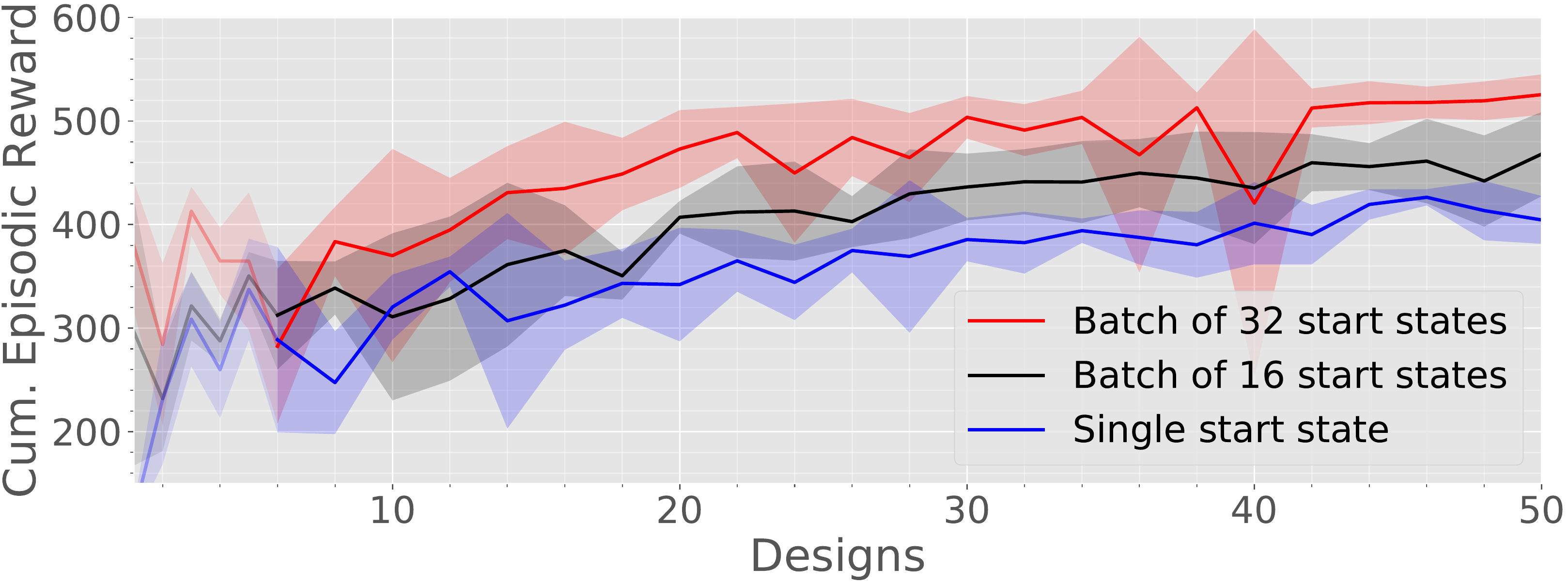}
    \caption{Evaluation of different batch sizes used in Eq. (\ref{Eq::objective}) on the Half-Cheetah task.}
    \label{Fig::Batch_sizes_in_q}
\end{figure}

\begin{figure}
    \centering
    \includegraphics[width=0.6\textwidth]{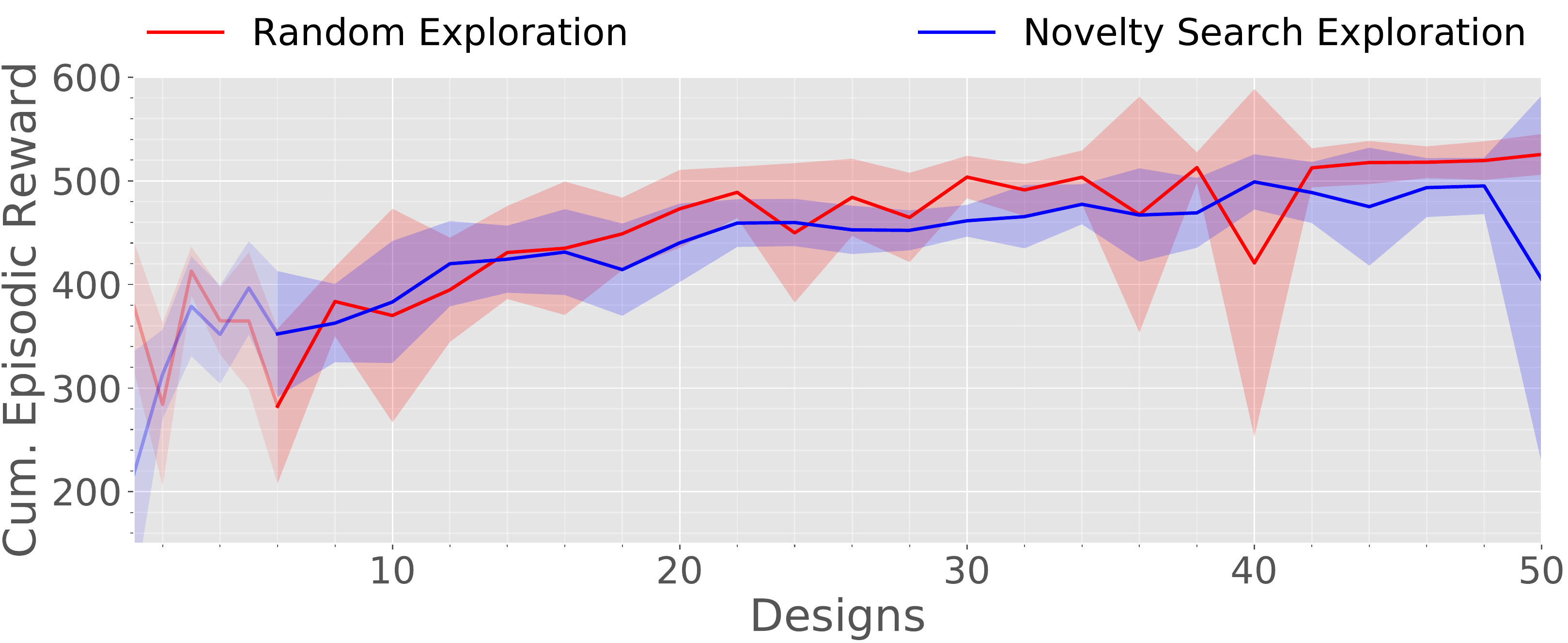}
    \caption{Evaluation of using novelty search or random selection as exploration mechanism during design optimization.}
    \label{Fig::novelty_vs_random}
\end{figure}
    
\begin{figure}
    \centering
    \includegraphics[width=0.6\textwidth]{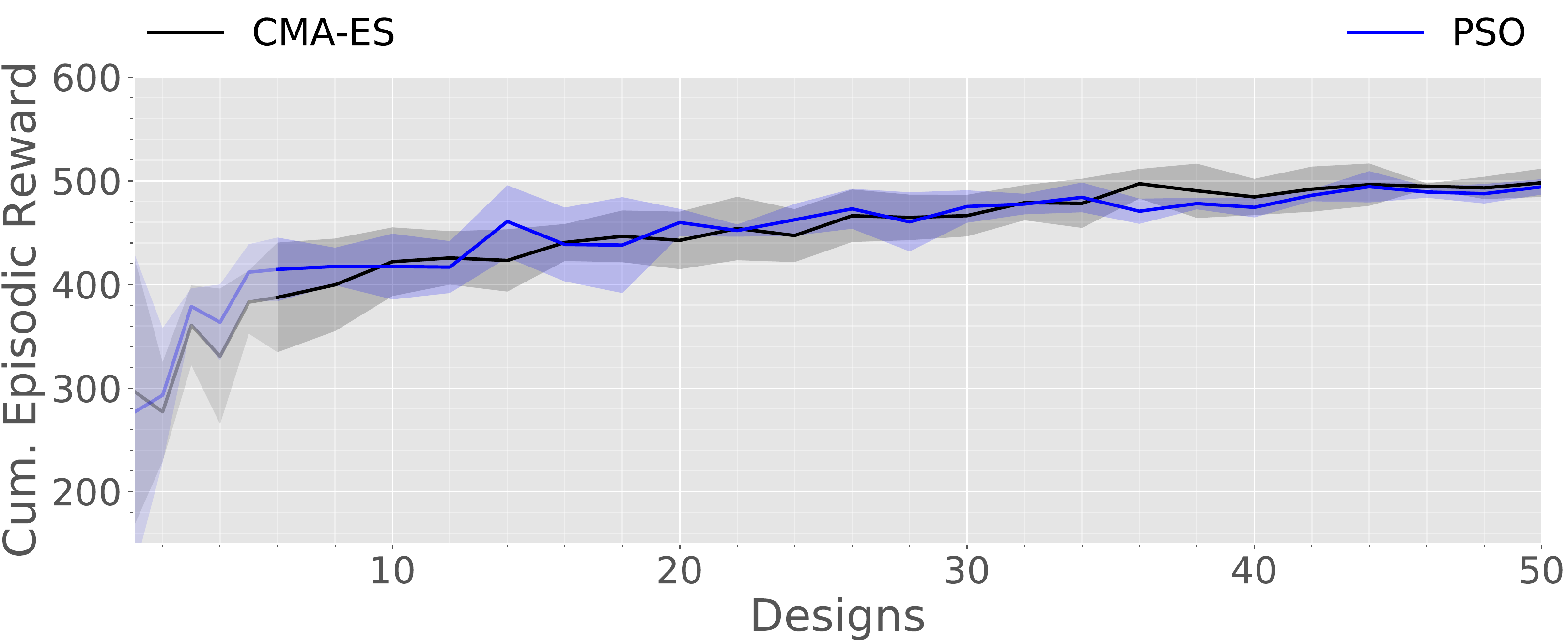}
    \caption{Comparison of CMA-ES and PSO when using rollouts from the simulator to optimize designs.}
    \label{Fig::Opt_and_Random::Optim}
\end{figure}

\begin{figure}
    \centering
    \includegraphics[width=0.6\textwidth]{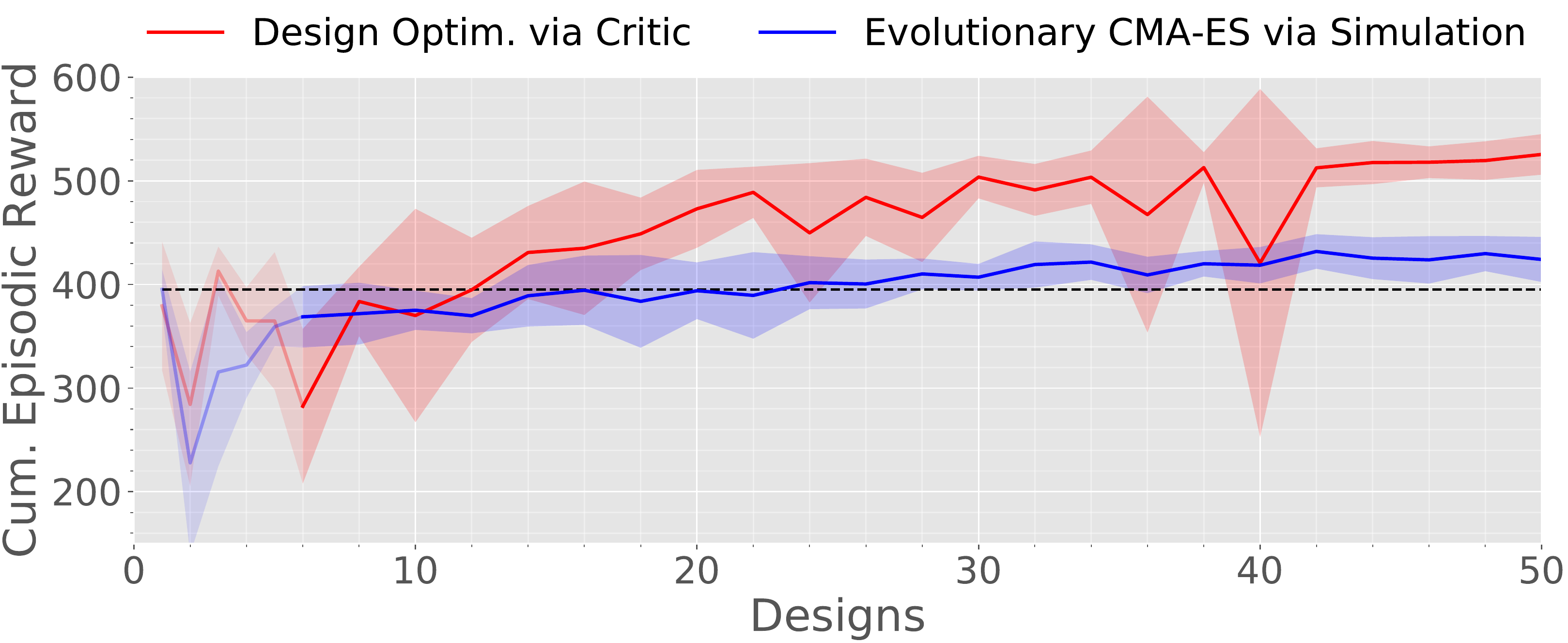}
    \caption{The plot shows, in blue, the use of CMA-ES used in the same manner as in \cite{ha2018reinforcement}. In each iteration lambda (here nine) design candidates are generated and evaluated in simulation. Then one iteration of CMA-ES is executed and the RL loop is executed on the best design found. Finally, a new population is generated and the next iteration of CMA-ES is executed. The dotted line shows the average best performance on the initial design. }
   \label{Fig::CMAES_EVO}
\end{figure}

\subsection{Using CMA-ES for Evolutionary Design Optimization}
As proposed in the work of David Ha \cite{ha2018reinforcement} we evaluated our approach against two approaches using CMA-ES (Fig. \ref{Fig::CMAES_EVO}) and OpenAI-ES (see main text) in an evolutionary manner for the optimization of the designs. 
For this experiment, we let CMA-ES create a population of design candidates and evaluated them in the simulator. 
We then executed exactly one update iteration of CMA-ES and used the best design found in the reinforcement learning loop. 
Figure \ref{Fig::CMAES_EVO} shows that this method is outperformed by the approach proposed in this paper. 
The proposed method uses the Q-function for design evaluations during the design optimization phase and executes a number of update iterations before selecting the best design for the reinforcement learning loop. 

\subsection{Design Exploration Strategies}
We alternate between design exploration and exploitation to increase the diversity of explored designs, improve generalization capabilities of the critic and avoid an early convergence to regions of the design space. 
Therefore, every time we find an optimal design during the design optimization process with the objective function (Eq. \ref{Eq::objective}) and conclude the subsequent reinforcement learning process, we next choose one design using the exploration strategy. 
To this end, we implemented two different approaches: sampling new designs 1) randomly, and 2) using Novelty search~\citep{lehman2008exploiting}. 
Novelty search is an exploration strategy in which the objective maximizes distance to the closest neighbours. 
The objective function is given by
\begin{equation}
    \max_{\design}\, \frac{1}{m} \sum_{\Tilde{\design} \in \text{NN}(\design, \Bar{\designspace})} \parallel \design - \Tilde{\design} \parallel_2,
\end{equation}
where the function $\text{NN}(\design, \Bar{\designspace})$ returns the $m$ nearest neighbors of a design $\design$ from the set $\Bar{\designspace}$ of chosen designs so far. 
This set includes only designs which were selected for evaluation in the real world or simulation, i.e., were handed over to the reinforcement learning algorithm as $\design_{\text{Opt}}$ (Fig. \ref{Fig::concepts::proposed}). 
Experiments showed that using novelty search for exploration did not yield an advantage over random selection of designs (Fig. \ref{Fig::novelty_vs_random}).

\subsection{Performance of Optimization Algorithms for Design Optimization}
Since we had to reduce the number of simulations considerably during the design optimization stage, we also evaluated the performance between Particle Swarm Optimization (PSO) and Covariance Matrix Adaptation-Evolution Strategy (CMA-ES). 
However, we could not find a significant difference in performance (Fig. \ref{Fig::Opt_and_Random::Optim}). 

\subsection{About the Use of Batches of Start States for the Evaluation of Design Candidates}
We evaluated the importance of evaluating the objective function (Eq.  \ref{Eq::objective}) over a batch of start states. 
Figure \ref{Fig::Batch_sizes_in_q} shows the use of a single start state $\klstate_0$, using a batch of 16 and 32 start states in the objective function presented in Eq.~\ref{Eq::objective}. 
The evaluation shows that averaging the objective function over a number of randomly drawn start states increases the performance of the proposed approach considerably. 
\begin{figure}[h]
    \centering
    \includegraphics[width=0.6\textwidth]{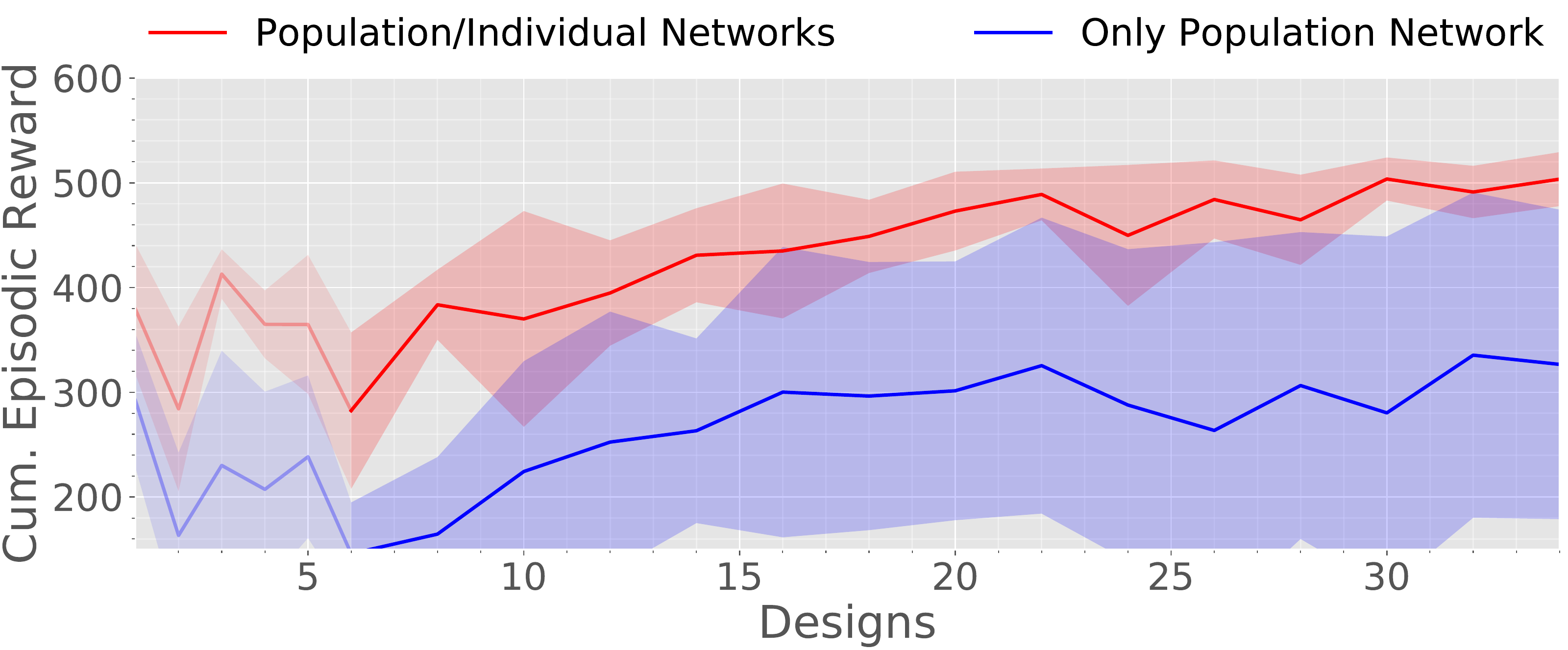}
    \caption{Using only a single \textit{population} network shows a worse performance than using the proposed combination of \textit{population} and \textit{individual} networks. This evaluation was performed on the Half-Cheetah task.}
    \label{fig:global_vs_local}
\end{figure}

\subsection{Evaluating the use of Population and Individual Networks}
In a preliminary evaluation we were able to confirm that the use of a single set of \textit{population} networks, instead of using a combination of \textit{population} and \textit{individual} networks, shows a decreased performance (Fig. \ref{fig:global_vs_local}). 
This shows that the ability of the \textit{individual} networks, to adapt quickly to the current design, is important for the overall performance of the proposed approach.

\clearpage

\end{document}